\documentclass[runningheads]{llncs}

\usepackage{eccv}

\usepackage{eccvabbrv}

\usepackage{graphicx}
\usepackage{booktabs}
\usepackage{array,multirow}
\usepackage[accsupp]{axessibility}  
\usepackage{mathtools, stmaryrd}
\DeclarePairedDelimiterX{\Iintv}[1]{\llbracket}{\rrbracket}{\iintvargs{#1}}
\NewDocumentCommand{\iintvargs}{>{\SplitArgument{1}{,}}m}
{\iintvargsaux#1} %
\NewDocumentCommand{\iintvargsaux}{mm} {#1\mkern1.5mu..\mkern1.5mu#2}

\usepackage{hyperref}

\usepackage{orcidlink}
\newcommand{\midtitle}[1]{
    \begin{center}
        \Large\textbf{#1}
    \end{center}
}

\begin{document}

\newcommand{\modelname}{\emph{\hbox{TransCAD}}}

\title{TransCAD: A Hierarchical Transformer for CAD Sequence Inference from Point Clouds} 

\titlerunning{\modelname: A Hierarchical Transformer for CAD Sequence}

\author{Elona Dupont\inst{1} \and
Kseniya Cherenkova\inst{1,2} \and
Dimitrios Mallis \inst{1} \and
Gleb Gusev \inst{2}
\and
Anis Kacem \inst{1}
\and
Djamila Aouada \inst{1}}
\authorrunning{E.~Dupont et al.}

\institute{SnT, University of Luxembourg, Luxembourg \and
Artec 3D, Luxembourg
}

\maketitle

\begin{abstract}
3D reverse engineering, in which a CAD model is inferred given a 3D scan of a physical object, is a research direction that offers many promising practical applications. This paper proposes \modelname, an end-to-end transformer-based architecture 
that predicts the CAD sequence from a point cloud. \modelname~leverages the structure of CAD sequences by using a hierarchical learning strategy. A loop refiner is also introduced to regress sketch primitive parameters. Rigorous experimentation on the DeepCAD~\cite{wu2021deepcad} and Fusion360~\cite{willis2021fusion} datasets show that \modelname~achieves state-of-the-art results. The result analysis is supported with a proposed metric for CAD sequence, the \textit{mean Average Precision of CAD Sequence}, that addresses the limitations of existing metrics. 
\end{abstract}


\section{Introduction} \label{sec:intro}

Practically every object encountered in daily life originates from a Computer-Aided Design (CAD), highlighting the fundamental role of CAD in industrial manufacturing processes. Currently, the dominant paradigm for CAD design is \textit{feature-based modelling}~\cite{xu2021zonegraph}. It allows the creation and manipulation of 3D models through a series of \textit{features}, individual elements or operations (holes, slots, fillets, \etc), that modify the geometry of a CAD model. The process is typically initiated with the design of \textit{planar sketches}, \ie collection of \textit{loops} composed of 2D curves, followed by a CAD operation (extrusion, revolution, etc.) that expands sketches into a 3D solid model. The final model is represented by the sequence of these CAD sketches and operations. \textit{Feature-based} modelling has been widely adopted, as it enables intuitive design alterations and seamless CAD software integration, making it essential for an iterative development of complex designs. The recent availability of large CAD model datasets, such as ABC~\cite{koch2019abc} and Fusion360~\cite{willis2021fusion}, has sparked significant interest in developing learning-based approaches for feature-based modelling. Recent efforts have been focused on deep generative modelling~\cite{wu2021deepcad, xu2023hnc, ren2022extrudenet, xu2022skexgen}, where large transformer-based networks are trained to create new CAD models or to automatically complete partial designs via autoregressive inference. While this research direction offers a lot of potential practical applications for CAD software integration, far less attention has been put to \textit{reverse engineering}. Feature-based reverse engineering emerges as a real-world application addressing the need to automatically replicate physical objects as CAD models.
Recovery of CAD design is facilitated by the acquisition of a point cloud or triangular mesh of a physical object scanned using commercial 3D sensors. 

Some existing reverse engineering approaches investigate the recovery of alternative CAD model representations like Constructive Solid Geometry~\cite{kania2020ucsg} (CSG) or Boundary-Representation (B-Rep)~\cite{lambourne2021brepnet, jayaraman2021uv, dupont2022cadops}. 
 Other methods tackle feature-based reverse engineering and predict implicit representations of sketches and CAD operations from point clouds~\cite{uy2022point2cyl,li2023secad}. Nevertheless, such approaches do not allow for seamless integration into CAD software and often require post-processing (\eg parametric curve fitting). To address these limitations, models capable of learning explicit CAD sequence of parametric sketches and operations from point clouds are needed. This can be enabled within a generative learning framework as in~\cite{wu2021deepcad}. In that work, an auto-encoder reconstructing CAD parametric sequences is proposed and the latent representation is used for generating novel CAD sequences. An extension for reverse engineering was proposed by replacing the CAD sequence encoder with a point cloud encoder trained to map point clouds to the latent representations. The main limitation of the above is the predefined latent space that cannot adapt to the variations present in real-world point clouds. This disconnection can cause the model to generalize poorly to unseen inputs, especially those with noise or irregularities that are present in 3D scans and that are not well-represented in the training data.

To that end, we propose \modelname~, a novel end-to-end trainable and single-stage hierarchical network for feature-based explicit CAD sequence reverse engineering from point clouds. Our network is hierarchical in the sense that it employs a two-tiered decoding process. Initially, a primary CAD sequence embedding is decoded, encapsulating high-level features of the design, that are then processed by secondary decoders, one dedicated to loop parameters and another to CAD operations. Each decoder specializes in a certain input, enabling a nuanced and precise recovery of CAD parameters. The decomposition of learned representations matches the decomposition inherent in the actual feature-based design process of conceptualizing a 3D model through distinct loop and operation steps. Moreover, \modelname~does not predict sketch primitive types explicitly as in~\cite{wu2021deepcad}; instead, we employ a unified primitive representation where types are determined solely by coordinates. Our formulation narrows the learning space by eliminating syntactically incorrect predictions and facilitates a seamless transition between different primitive types. Additionally, it allows for a cascaded parameter refining that further enhances model performance.

Another focus of this work is the evaluation of parametric CAD sequence. We identify several limitations of the existing evaluation framework used by~\cite{wu2021deepcad, multicad} and suggest a suitable metric for CAD sequence similarity based on mean average precision, computed in the unquantized parametric space.

\vspace{0.1cm}
\noindent \textbf{Contributions:} In summary our contributions are the following:
\begin{enumerate}
    \item We propose \modelname, a novel hierarchical architecture for feature-based reverse engineering. Our model is single-stage and end-to-end trainable. \modelname~allows for a compact CAD sequence representation that does not include categorical types and enables cascaded coordinate refinement. 
    \item We identify several limitations of the existing evaluation framework for feature-based reverse engineering and propose a new evaluation metric framework to ensure fair comparison among diverse network architectures.
    \item Our model surpasses the performance of recent generative-based approaches while also bridging the gap to real-world applications by exhibiting robustness to perturbed point clouds.
\end{enumerate}

\noindent \textbf{Paper Organization:} The rest of the paper is organized as follows. Section \ref{sec:related_works} reviews related works. Section~\ref{sec:formulation} formulates the problem of feature-based CAD reverse-engineering. The proposed \modelname~is described in Section~\ref{sec:network_architecture}. Discussion on the current evaluation framework and suggested extension is introduced in Section~\ref{sec:evaluation_framework}. An experimental validation of the proposed network is provided in Section~\ref{sec:experiment}. Finally, conclusions are given in Section~\ref{sec:conclusion}.

\section{Related Works}
\label{sec:related_works}

\noindent \textbf{Generative Models for CAD:} The advent of large-scale 3D shape datasets \cite{chang2015shapenet, lim2013parsing, koch2019abc, willis2021fusion}, combined with the significant progress for generative models in vision \cite{Goodfellow2022GenerativeAN, Ho2020DenoisingDP, Karras_2020_CVPR, Chang_2022_CVPR}, has sparked interest in the generation of 3D shapes. Existing methods have been proposed for various 3D representations, including point clouds\cite{achlioptas2018learning, yang2019pointflow, cai2020learning}, 3D meshes~\cite{wang2018pixel2mesh, nash2020polygen}, voxel grids~\cite{wu2016learning, li2017grass}, and signed distance functions~\cite{chen2019learning, Wu_2020_CVPR}. This work focuses on CAD model generation, which compared to the above is parametric and directly editable in CAD software. A line of work explores the generation of the Boundary-Representation (B-Rep), a collection of parametric surfaces connected via a structured topological graph. SolidGen~\cite{jayaraman2022solidgen} considers B-Rep synthesis based on transformers and two-level pointer networks. BrepGen~\cite{xu2024brepgen} represents a B-Rep via a fixed tree of latent geometry representations that can be generated by a diffusion model. Feature-based CAD generation has also been recently explored. Most relevant to our work is DeepCAD~\cite{wu2021deepcad}, a non-autoregressive generative model capable of synthesizing novel CAD sequences based on a transformer auto-encoder architecture. In~\cite{xu2023hnc, xu2022skexgen} the authors also follow autoregressive strategies. HNC~\cite{xu2023hnc} uses a hierarchical model based on high-level concepts and a code tree for CAD model generation and auto-completion. Similarly, in SkexGen~\cite{xu2022skexgen} a transformer architecture is used to generate CAD models in the sketch-extrude format by encoding the topology and geometry using different codebooks. 
All the aforementioned works are oriented around 3D shape generation and either do not address the reverse-engineering task or address it via adaptation of generative modelling leading to suboptimal performance.

\vspace{0.2cm}
\noindent \textbf{CAD Reverse Engineering:} Reverse engineering is a well-studied problem with a substantial research effort directed towards predicting geometric features of CAD models, by analyzing the corresponding point clouds. Parametric fitting techniques infer the parametarization of edges~\cite{liu2021pc2wf, cherenkova2023sepicnet, wang2020pienet, yu2018ecnet, sharma2020parsenet, zhu2023nerve} and surfaces~\cite{li2019spfn,guo2022complexgen}. Various attributes of the B-Rep and CAD operations are recovered from 3D scans in~\cite{mallis2023sharp}. CADOps-Net~\cite{dupont2022cadops} recovers 2D sketches from faces segmented into their CAD operation steps. Reasoning about a CAD model via properties discovered by parametric fitting offers insights solely into its end-state, without considering the sequential CAD design process intrinsic to feature-based modeling.

A step closer to CAD reverse engineering, a line of work explores the reconstruction of a point cloud into Constructive Solid Geometry (CSG)~\cite{friedrich2019optimizing, kania2020ucsg, yu2024d}, a modelling technique that uses boolean operations to combine primitives into 3D models. Point2Cyl~\cite{uy2022point2cyl}, on the other hand, predicts extrusion cylinders given a point cloud, but requires user input to combine cylinders. SECAD-Net~\cite{li2023secad} and ExtrudeNet~\cite{ren2022extrudenet} use a self-supervised learning strategy to recover CAD sequences in the form of implicit representations given voxels and point clouds, respectively. In contrast to feature-based modelling, 3D representations produced by these methods (CSG, extrusion cylinders, \etc) have limited compatibility with modern CAD software workflows. The authors in~\cite{lambourne2022reconstructing} learn sketch-extrude sequences conditioned on a voxel input, however, the model relies on strong data priors and is limited to predefined extrusion combinations.

Closer to our work is DeepCAD~\cite{wu2021deepcad} and subsequent MultiCAD~\cite{multicad}. Even though DeepCAD~\cite{wu2021deepcad} proposes a non-autoregressive generative framework for feature-based CAD, authors explore further conditioning on input point clouds. Taking a similar direction, MultiCAD~\cite{multicad} proposes a two-stage multimodal contrastive learning strategy of both point clouds and CAD sequences. The two aforementioned methods opt for separate stage learning for point clouds and CAD sequences. Concurrent to our work, the autoregressive strategy in~\cite{khan2024cad} and the multimodal diffusion based approach in~\cite{ma2024draw} attempt to solve the point cloud to CAD sequence problem. To our knowledge \modelname~is the first non-autoregressive single-stage architecture for feature-based reverse engineering.

\section{Problem Statement}
\label{sec:formulation}

A CAD model $\mathbf C \in \mathcal{C}$ is constructed in a sequence of construction steps. Each step can be seen as a 2D parametric \textit{sketch} $\mathbf s \in \mathcal{S}$ (\eg set of lines, arcs, \etc) followed by a CAD \textit{operation} $\mathbf o \in \mathcal{O}$ (\eg extrusion, revolution, \etc)~\cite{xu2022skexgen,wu2021deepcad}. Here, $\mathcal C$ is the set of all possible CAD models, $\mathcal S$ and $\mathcal O$ represent the sets of possible CAD sketches and operations, respectively. CAD models constructed exclusively from the \textit{extrusion} operation type are considered in this work. Extrusion $\mathbf{e}\in\mathcal{E}$, where $\mathcal{E}$ denotes the set of possible extrusions, is the most common operation and enables the description of a wide range of CAD models~\cite{wu2021deepcad,xu2022skexgen,xu2023hnc}. \modelname~aims at learning how to predict the sequence of CAD construction steps from an input point cloud. Formally, given a point cloud $\mathbf{P}~=~[\mathbf p_1,\dots,\mathbf p_n]~\in~\mathbb{R}^{n\times
3}$, where $\mathbf p_i~=~[x_i,y_i,z_i]$ denotes the 3D coordinates of the point $i$ and $n$ the number of points, the objective of \modelname~is to learn the mapping 

\begin{equation}
\begin{split}
\mathbf{\Phi} :  \mathbb{R}^{n\times3}& \rightarrow \mathcal C \ , \\
\mathbf{\Phi}(\mathbf{P})= \{&\mathbf{s}_l,\mathbf{e}_l\}_{l=1}^L \ ,
\end{split}
\end{equation}
\noindent where $L$ denotes the length of the CAD sequence. In what follows, the proposed formulations of sketches and extrusions are described. 
\begin{figure}[t]
\begin{center}
\setlength{\belowcaptionskip}{-0.6cm}
\includegraphics[width=\linewidth]{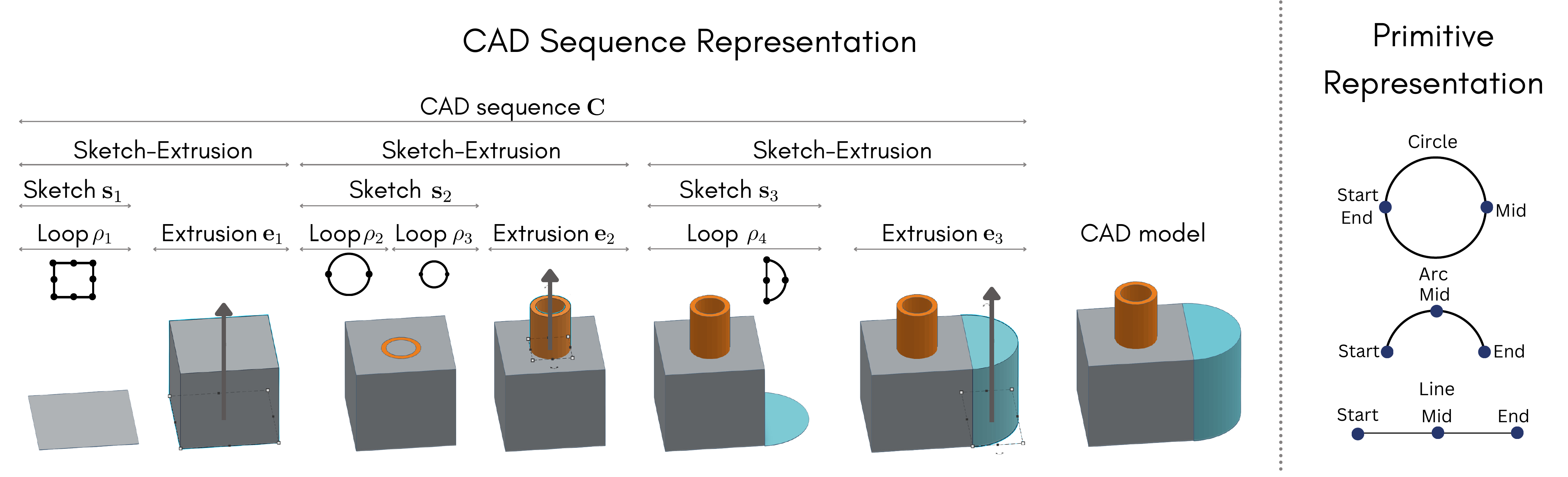}
\caption{The sequential process of CAD modeling. A CAD sequence $\mathbf C$ can be decomposed into a hierarchical structure. The highest conceptual level is a sequence of sketch $\mathbf s$ and extrusion $\mathbf e$. A sketch can be made of one or more loops $\mathbf \rho$. Each loop can be further decomposed into loop primitives, circle, arc and line. Each loop primitive can be described by a fixed number of parameters as shown on the right panel.}
\label{fig:cad_seq}

\end{center}
\end{figure}

\vspace{0.2cm}
\noindent\textbf{CAD Sketch and Extrusion Formulation:} The proposed formulations for sketch and extrusion steps are inspired by~\cite{wu2021deepcad, xu2022skexgen}.
A sketch $\mathbf s$ is composed of one or more \textit{loops} (see left panel of Fig.~\ref{fig:cad_seq}). Each loop $\{\mathbf{\rho}_j\}_{j=1}^{L_{\rho}}$, where $L_{\rho}$, denoting the number of loops, consists of one primitive (\ie circle) or a combination of primitives (\ie lines and arcs). In contrast to~\cite{wu2021deepcad} which specifies the type of primitives in their representation, the proposed primitive representation is type-agnostic (see right panel of Fig.~\ref{fig:cad_seq}). In particular, each primitive $\delta$ is represented by three 2D coordinates of start, mid, and end points, \ie  $\delta~=~[(x_{start},y_{start}), (x_{mid},y_{mid}), (x_{end},y_{end})]~\in~\mathbb R^6$. This representation has the advantage that the type of primitive can be deduced from the configurations of the points, hence reducing the search space and facilitating the transition between different types during training. In practice, the mid point of a line is replaced by a dummy value. 

As in~\cite{wu2021deepcad}, we ensure that the loops are always closed by using the end point of a primitive as the start point of the next one. Further, a similar to~\cite{wu2021deepcad} quantization is considered to reduce the parameters search space. As a result, a loop of $n_p$ primitives $\mathbf{\rho}_j~\in~\mathbb R^{6 \times n_p}$  is considered as a quantized representation $\mathbf{\rho}^{\star}_j \in \Iintv{0,d_{q}}^{6\times n_p}$, where $d_q$ denotes the quantization interval. As for extrusion, similarly to ~\cite{wu2021deepcad}, a quantized representation $\mathbf{e}^{\star}_j~\in~\Iintv{0,d_{q}}^{11}$ is considered to represent the sketch plane/scale and extrusion type/distances. Note that $\mathbf{e}^{\star}~\in~\Iintv{0,d_{q}}^{11\times L_e}$ and $\mathbf{\rho}^{\star}~\in \Iintv{0,d_{q}}^{6\times n_p \times L_{\rho}}$ will be used in the following to denote sequences of quantized extrusions and loops, respectively. Here, $L_{\rho}$ and $L_e$ denote the length of loop and extrusion sequences, respectively.

\section{Hierarchical CAD Sequence Learning from Point Clouds}
\label{sec:network_architecture}
\begin{figure}[t]
\begin{center}
\setlength{\belowcaptionskip}{-0.6cm}
\includegraphics[width=\linewidth]{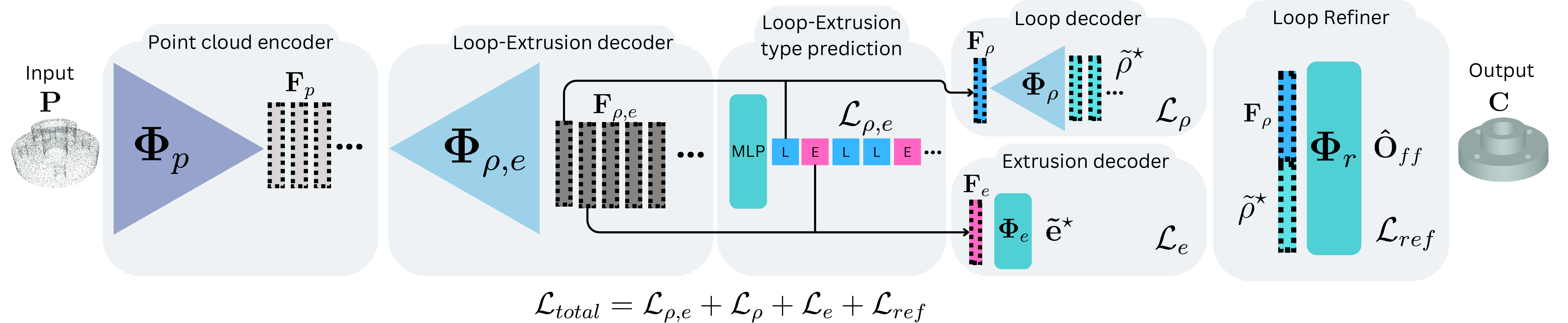} 
\caption{\modelname~model architecture. \modelname~is a hierarchical network composed of the following components: a point cloud encoder, a loop-extrusion decoder that predicts a high-level sequence which is then decoded by a loop decoder and an extrusion decoder. The predicted quantized loop parameters are then corrected by a loop refiner.}
\label{fig:model}
\end{center}
\end{figure}

\modelname~non-autoregessively learns to predict a CAD sequence from an input point cloud in the format described in Section~\ref{sec:formulation}. First, the point cloud is encoded into point features using a standard point cloud encoder. In order to facilitate the learning of CAD sequences, a hierarchical CAD sequence decoding is proposed. In particular, a high-level sequence of embedding corresponding to the loop and extrusion steps is learned. Those embedding are then fed to either a loop or extrusion decoder based on predicted type to learn the loop and extrusion parameters. 
Finally, the predicted loop parameters are further refined using the actual unquantized loop parameters (ground truth). The overall model architecture is depicted in Fig.~\ref{fig:model} and the different components are described below.

\subsection{Point Cloud Encoder}

The point cloud encoder $\mathbf{\Phi}_p$ consists of $4$ layers of PointNet++~\cite{qi2017pointnet++} and operates on an input point cloud $\mathbf{P}$.  
It outputs per-point features encoding local neighborhood information, $\mathbf{F}_p~=~[\mathbf{f}_p^1\dots\mathbf{f}_p^n]~\in~\mathbb R^{n\times d_p}$, that encode local neighborhood information, $d_p$ denotes the dimension of the features. Note that point normals of $\mathbf{P}$ are estimated using~\cite{open3d_estimate_normals} and are provided as input to $\mathbf{\Phi}_p$ along with its 3D coordinates.

\subsection{Loop-Extrusion Decoder}

 The main objective of the loop-extrusion decoder $\mathbf{\Phi}_{\rho,e}$ is to learn a high-level sequence of embedding $\mathbf{F}_{\rho,e}~=~[\mathbf{f}_{\rho,e}^1,\dots,\mathbf{f}_{\rho,e}^{L_{\rho,e}}]~\in~\mathbb{R}^{d_{z} \times L_{\rho,e}}$ corresponding to loops and extrusions from the point cloud features $\mathbf{F}_p$. Here, $d_z$ and $L_{\rho,e}$ denote the embedding dimension and the length of the sequence, respectively. The decoder $\mathbf{\Phi}_{\rho,e}$ is composed of multi-head transformer-based blocks~\cite{attention}. In the first block, learned constant embedding $\mathbf{F}_c~\in~\mathbb{R}^{d_z \times L_{\rho,e}}$ undergo a self-attention operation~\cite{attention} and the resulting representation cross-attends to the point cloud features $\mathbf{F}_p$ to produce loop extrusion embedding for the first block $\mathbf{F}_{\rho,e}^1~\in~\mathbb{R}^{d_z \times L_{\rho,e}}$ as follows, 
\begin{equation}
    \mathbf{F}_{\rho,e}^1 = \texttt{CA}(\texttt{SA}(\mathbf{F}_c),\mathbf{F}_p) \ ,
     \label{eq:sa-ca-1st}
\end{equation}

\noindent where $\texttt{SA(.)}$ and $\texttt{CA}( .,. )$ denote the self and cross attention operators~\cite{attention}, respectively. The same self and cross attention operations are conducted in the subsequent blocks by feeding the output of each block as input to the next one,

\begin{equation}
    \mathbf{F}_{\rho,e}^b = \texttt{CA}(\texttt{SA}(\mathbf{F}_{b-1}),\mathbf{F}_p) \ ,
    \label{eq:sa-ca-blocks}
\end{equation}

\noindent yielding the final sequence of embedding $\mathbf{F}_{\rho,e}$ after the last block. In order to ensure that each element $\mathbf{f}_{\rho,e}^i$ in the sequence embedding $\mathbf{F}_{\rho,e}$ corresponds to the right type (\ie loop, extrusion, or end of sequence), a $3$ layer $\texttt{MLP}$ followed by $\texttt{softmax}$~that operates on each $\mathbf{f}_{\rho,e}^i$ and predicts its type is introduced. A cross-entropy loss, $\mathcal{L}_{\rho,e}$, is computed between the predicted and ground truth types to supervise the learning of $\mathbf{F}_{\rho,e}$. Note that the loop-extrusion decoder is solely used to obtain a high-level sequence of loop and extrusion embedding. These embedding can then be decoded through either a loop decoder or an extrusion decoder to obtain their parameters. At training time, the ground truth type labels are used to identify which decoder should be used for each embedding, while at inference time the predicted types are used. The identification of loop and extrusion types results into separate loop embedding $\mathbf{F}_{\rho}~\in~\mathbb{R}^{d_z \times L_{\rho}}$ and extrusion embedding $\mathbf{F}_{e}~\in~\mathbb{R}^{d_z \times L_e}$ by splitting $\mathbf{F}_{\rho,e}$ according to loop and extrusion types.

\subsection{Loop and Extrusion Parametrization}

After obtaining the representation and the type of loop and extrusion steps, the parameters of both loops and extrusions are decoded from these representations using separate decoders. 

\vspace{0.2cm}
\noindent \textbf{Extrusion Decoder:} As mentioned in Section~\ref{sec:formulation}, the extrusion sequence is described by a sequence of $11$ quantized parameters $\mathbf{e}^{\star}~\in~\Iintv{0,d_{q}}^{11\times L_e}$. In order to obtain these parameters from the extrusion sequence embedding $\mathbf{F}_{e}$, an extrusion decoder $\mathbf{\Phi}_e$ consisting of $3$ $\texttt{MLP}$ layers followed by $\texttt{softmax}$ is used. The predicted probabilities of the extrusion sequence parameters $\mathbf{\Tilde{e}}^{\star}~\in~[\mathbf{0},\mathbf{1}]^{11 \times d_q \times L_e}$ are compared to the ground truth one-hot-encoded parameters in $\mathbf{e}^{\star}$ using a cross-entropy loss, $\mathcal{L}_{e}$. 

\vspace{0.2cm}
\noindent \textbf{Loop Decoder:} Similarly to the extrusion decoder, the loop decoder $\mathbf{\Phi}_{\rho}$ predicts the quantized parameters of the loop sequence $\mathbf{\rho}^{\star}~\in \Iintv{0,d_{q}}^{6\times n_p \times L_{\rho}}$ as explained in Section~\ref{sec:formulation}. Nevertheless, $4$ layers of multi-head transformer blocks are employed instead of simple $\texttt{MLP}$ layers. This is due to the sequential nature of loop decoding in contrast to extrusions. Note that a similar strategy as loop-extrusion decoder is opted for the transformer block of loop decoder. The first block performs self-attention on learned constant embedding of loops  $\mathbf{F}_c^{\rho}~\in~\mathbb{R}^{d_{z} \times n_pL_{\rho}}$ and the result cross-attends to loop embedding $\mathbf{F}_{\rho}$ as in Eq.(\ref{eq:sa-ca-1st}). The same self and cross attention operations in Eq.(\ref{eq:sa-ca-blocks}) are conducted in subsequent blocks to yield a final representation at the last block $\mathbf{F}_{\rho}^b~\in~\mathbb R^{d_{z} \times n_pL_{\rho}}$. A linear layer followed by $\texttt{softmax}$ is used to obtain predicted probabilities for the loop sequence parameters $\Tilde{\mathbf{\rho}}^{\star}~\in~[\mathbf{0},\mathbf{1}]^{ 6 \times d_q \times n_pL_{\rho}}$ which are compared to ground truth one-hot-encoded loop parameters of $\mathbf{\rho}^{\star}$ using a cross-entropy loss, $\mathcal{L}_{\rho}$. 

\vspace{0.2cm}
\noindent \textbf{Loop Refiner:} As in many transformer-based architectures~\cite{wu2021deepcad,xu2022skexgen,carlier2020deepsvg,attention}, the quantization of loop parameters helps to reduce the search space and facilitates the learning. However, it has been observed in our case that it can lead to accumulation of quantization approximation errors. To overcome this issue, unquantized ground truth loop parameters are leveraged. In particular, a loop refiner $\mathbf{\Phi}_r$ composed of a $4$ layer $\texttt{MLP}$ is introduced. This refiner takes as input a concatenation of loop embedding $\mathbf{F}_{\rho}$ and their corresponding predicted parameter probabilities $\Tilde{\mathbf{\rho}}^{\star}$. It attempts to predict the offset $\mathbf{{\hat{O}}}_{ff}~\in~\mathbb{R}^{6\times n_pL_{\rho}}$ between the predicted quantized loop parameters  $\hat{\mathbf{\rho}}^{\star}~\in~\Iintv{0,d_{q}}^{ 6 \times n_pL_{\rho}}$ and the unquantized ground truth loop parameters $\mathbf{\rho}^{\star}~\in~\mathbb{R}^{6\times n_pL_{\rho}}$. An MSE loss, $\mathcal{L}_r$, is computed between the predicted offset $\mathbf{{\hat{O}}}_{ff}$ and the one given by $\mathbf{{{O}}}_{ff} = \mathbf{\rho}^{\star} - \mathbf{\rho}$, to supervise the refiner and the rest of the network. Once the offset is predicted, it is added to the predicted quantized loop parameters yielding unquantized predicted loop parameters as follows $\mathbf{\hat{\rho}} =  \mathbf{\hat{\rho}}^{\star} + \mathbf{{\hat{O}}}_{ff}$. 

\vspace{0.2cm}
\noindent \textbf{Total Loss:} \modelname~is an end-to-end network with a training objective guided by the sum of the individual losses, $\mathcal{L}_{total} = \mathcal{L}_{\rho,e} + \mathcal{L}_{\rho} + \mathcal{L}_{e} + \mathcal{L}_{r}$.

\section{Proposed Evaluation}
\label{sec:evaluation_framework}

In this section, we first outline the limitations of existing evaluation methods in CAD sequence. Then, our new proposed evaluation metric framework for assessing the performance of CAD sequence inference from point clouds is described.

\vspace{0.2cm}
\noindent \textbf{DeepCAD\cite{wu2021deepcad} Evaluation for Feature-based Reverse Engineering:} An evaluation framework for CAD sequence was originally introduced in~\cite{wu2021deepcad} and later used in~\cite{multicad}. This framework includes both accuracy for assessing the fidelity of the predicted sequence and Chamfer Distance ($CD$) to measure the quality of the recovered 3D geometry.
Accuracy is assessed using two metrics, specifically \textit{Command Type Accuracy} ($ACC_{cmd}$) and \textit{Parameter Accuracy} ($ACC_{param}$) defined by 
\begin{align}
    ACC_{cmd} &= \frac{1}{N_c} \sum_{i=1}^{N_c} \mathbb{I}[t_i = \hat{t_i}] \ , \label{eq:cmd_accuracy} \\
    ACC_{param} &= \frac{1}{K} \sum_{i=1}^{N_c} \sum_{j=1}^{|\hat{\boldsymbol{p}}_{i}|} \mathbb{I}\left[|\boldsymbol{p}_{i,j} - \hat{\boldsymbol{p}}_{i,j}|<\eta\right]\mathbb{I}[t_i = \hat{t_i}] \ , \label{eq:param_accuracy}
\end{align}
\noindent where $t_i$ and $\hat{t_i}$ are the ground truth and predicted command types (for commands representing primitives and extrusions), $\boldsymbol{p}_{i,j}$ and $\hat{\boldsymbol{p}}_{i,j}$ are ground truth and predicted command parameters, $N_c$ denotes the total number of CAD commands and $\mathbb{I[.]}$ is the indicator function. $K = \sum_{i=1}^{N_c} \mathbb{I}[t_i = \hat{t_i}]|\boldsymbol{p}_i|$ is the total number of parameters of the correctly recovered commands and $\eta$ is a tolerance threshold. 
The 3D geometry is evaluated with Chamfer Distance ($CD$) computed by sampling $2000$ points on the ground truth and predicted shapes. 

\begin{figure}[ht]
\begin{center}
\setlength{\belowcaptionskip}{-0.6cm}
\includegraphics[width=0.8\linewidth]{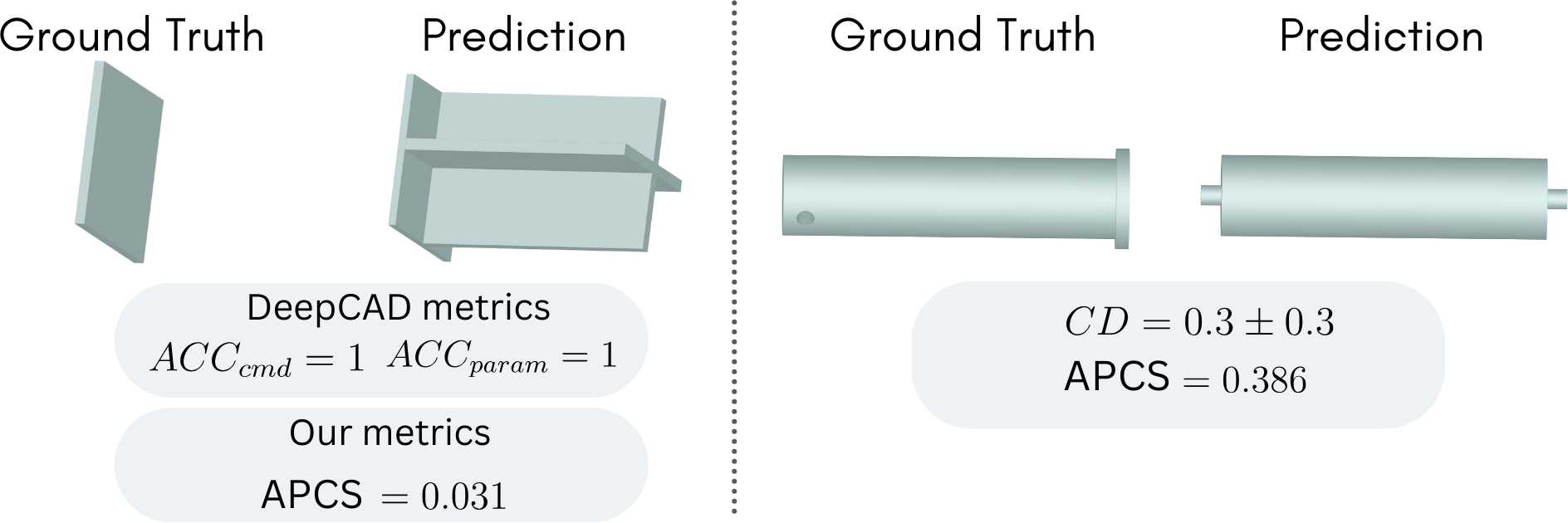}
\end{center}
 \caption{Two examples outlining the limitations of existing evaluation metrics. Left panel: the ground truth sequence (one sketch-extrusion) is a correct subset of the predicted sequence (three sketch-extrusions). The DeepCAD~\cite{wu2021deepcad} metrics result in an accuracy of $1$ for both commands and parameters. On the other hand, our proposed metric takes into account the over predicted sequence elements and the APCS is 0.031. The right panel showcases the limitations of the $CD$ as a similarity measure. While the ground truth and predicted shapes are both composed of three extruded circle sketches, they are different in shape. However, the $CD$ between the two shapes falls within the uncertainty range of $\pm 0.3$. Note that the uncertainty in the $CD$ measurement is estimated by taking the average $CD$ between all the test samples and themselves.}
\label{fig:metrics}
\end{figure}

\vspace{0.2cm}
\noindent \textit{Limitations:} We identify the following limitations of the aforementioned evaluation. \textbf{(1)} The proposed $ACC_{cmd}$ overlooks the possibility of over-prediction in the CAD sequence. As indicated in Eq.(\ref{eq:cmd_accuracy}), the computation of this metric sums across the set of ground truth CAD commands $N_c$. A predicted sequence could erroneously include extra loop-extrusion operations and still achieve a full score, as exemplified on the left panel of Fig.~\ref{fig:metrics}. \textbf{(2)} The evaluation of $ACC_{param}$ is conducted solely on the subset of $K$ accurately identified commands, thus introducing a trade-off between $ACC_{param}$ and $ACC_{cmd}$. This interdependence complicates the interpretation of results. 
\textbf{(3)} Assessment of parameter quality via $ACC_{param}$ solely in terms of accuracy is failing to distinguish between the magnitudes of errors. A parameter inaccurately placed in an adjacent quantization bucket incurs the same penalty as one with a larger deviation, despite potentially minor implications on the CAD model's final geometry. These limitations cannot be entirely mitigated by complementing CAD command accuracies with the chamfer distance ($CD$) metric. While $CD$ is a valuable assessment of shape similarity, it does not address the core objective of reverse engineering: to accurately recover the designer's original CAD sequence. Two CAD models might be close in terms of $CD$ yet possess vastly different CAD construction steps (see right panel of Fig.~\ref{fig:metrics}).

\vspace{0.2cm}
\noindent\textbf{Proposed Evaluation Framework:} To overcome the identified challenges, we introduce the \textit{mean Average Precision of CAD Sequence} (APCS), a novel evaluation metric tailored for feature-based reverse engineering. APCS adopts the concept of Average Precision (AP) commonly used in other tasks, to quantify the similarity between predicted and ground truth CAD sequences. We introduce the \textit{CAD Sequence Similarity Score} (CSSS) that can be computed between predicted and ground truth CAD sequences as follows

\begin{equation}
 \resizebox{0.93\linewidth}{!}{
    $CSSS(\hat{\mathbf{C}}, \mathbf{C})=  \frac{1}{2N_\delta}\sum_{j=1}^{N^\rho} \sum_{i=1}^{N_j^\rho } \left[S(\hat{\delta}_{j,i}, \delta_{j,i})\cdot\mathbb{I}[t_{yp}(\hat{\delta}_{j,i}) = t_{yp}(\delta_{j,i}) ]\right]+ \frac{1}{2N^e}\sum_{j=1}^{N^e} S(\hat{\boldsymbol{e}}_j, \boldsymbol{e}_j) \ , 
    $ } 
\end{equation}

\noindent where $t_{yp}(.)$ is a function that determines the type of each primitive $\delta$ (arc, line, \etc), and $S(\hat{\boldsymbol{p}},\boldsymbol{p}) = e^{-k||\hat{\boldsymbol{p}} - \boldsymbol{p}||}$ is a scoring function with $S(\hat{\boldsymbol{p}},\boldsymbol{p}) \in [0,1]$ with $1$ assigned when predicted parameterization is identical to the ground truth. We define $N_j^\rho = \max{(|\rho_j|,|\hat{\rho}_j|)}$ where $|.|$ denotes set cardinality, $N^\rho = \max{(L_\rho,\hat{L_\rho})}$ where $L_\rho$ is the number of predicted primitives and $\hat{L}_\rho$ is the number of ground truth primitives for loop $\rho_j$ and $N_\delta=\sum_{j=1}^{N^\rho} \max{(|\rho_j|,|\hat{\rho}_j|)}$. Finally, $N^e~=~\max{(L_e,\hat{L_e})}$ where $L_e$ is the number of predicted extrusions and $\hat{L}_e$ is the number of ground truth extrusions. The proposed CSSS metric evaluates both the operation type and parameter prediction. It assigns a score of $0$ to loops with incorrectly predicted types, which gradually increases to 1 as parameter prediction improves. Assessment is conducted on the unquantized parameter space and calculates the score based on the maximum count of either predicted or ground truth primitives. This approach ensures that both over and under predicted sequences are penalized equally. We aggregate CSSS scores across various thresholds to derive the \textit{mean Average Precision of CAD Sequences} (APCS).
Furthermore, the median $CD$ is used to measure shape similarity as in~\cite{wu2021deepcad} with the difference that it is evaluated on 4096 points instead of 2000 in order to decrease the uncertainty in the $CD$ measurement. All the reported $CD$ measurements in this work are multiplied by $10^3$. Moreover, the ratio of predictions that cannot be reconstructed using~\cite{opencascade} is reported as the invalidity ratio, IR.

\section{Experiments}
\label{sec:experiment}

In this section, the experimental setup is first presented. Then, qualitative and quantitative results are analyzed. Afterwards, the components of \modelname~are ablated. Finally, the limitations of our model are outlined.

\subsection{Experimental Setup} \label{sec:exp_setup}

\noindent 
\textbf{Dataset}: For training and evaluation, the DeepCAD dataset~\cite{wu2021deepcad} is used. The sketch extrusion sequences of the CAD models are processed in quantized ($8$ bits) and unquantized space. The size of the train, validation and test sets are $140 \,294$, $7\,773$, and $7\,036$ CAD models, respectively. Moreover, cross-dataset evaluation is conducted on the Fusion360 dataset~\cite{willis2021fusion} that contains $6\,794$ samples.

\vspace{0.2cm}
\noindent
\textbf{Training Details}: The network is trained for $100$ epochs with a batch size of $72$ and an Adam optimizer is employed with a learning rate of $0.001$ and a linear warm-up period of $2\,000$ steps as in~\cite{wu2021deepcad}. The training is conducted on an NVIDIA RTX $A6000$ GPU. The input point clouds are extracted using~\cite{opencascade} and are made of $n=4096$ points. The dimension of the point features $d_p$ is set to $16$ and of loop-extrusion features $d_z$ to $256$.

\vspace{0.2cm}
\noindent
\textbf{Baselines}: In order to evaluate the performance of \modelname, two state-of-the-art methods, MultiCAD~\cite{multicad} and DeepCAD~\cite{wu2021deepcad}, and a retrieval baseline are used. As the code for MultiCAD~\cite{multicad} is not available, we report the results from the original paper. DeepCAD~\cite{wu2021deepcad} is retrained with the same parameters and procedure as outlined in the original paper. One of the known limitations of the DeepCAD dataset is that it contains duplicate models~\cite{xu2022skexgen}. While the works in~\cite{xu2022skexgen} proposed a method to remove duplicate models that contain exactly the same CAD sequence, we find that this method does not remove all the duplicates as some models can have the same geometry but are constructed through a slightly different sequence of sketch extrusion operations (see supplementary material for more details). In order to address this issue, we propose a retrieval baseline. The retrieval baseline uses the point cloud encoder from a trained DeepCAD~\cite{wu2021deepcad} to identify the closest latent vector from the train set for each test sample. As a result, the solution is always a train set CAD sequence.

\subsection{Experimental Results}  \label{sec:exp_results}

\begin{figure}[ht]
\begin{center}
\includegraphics[width=\linewidth]{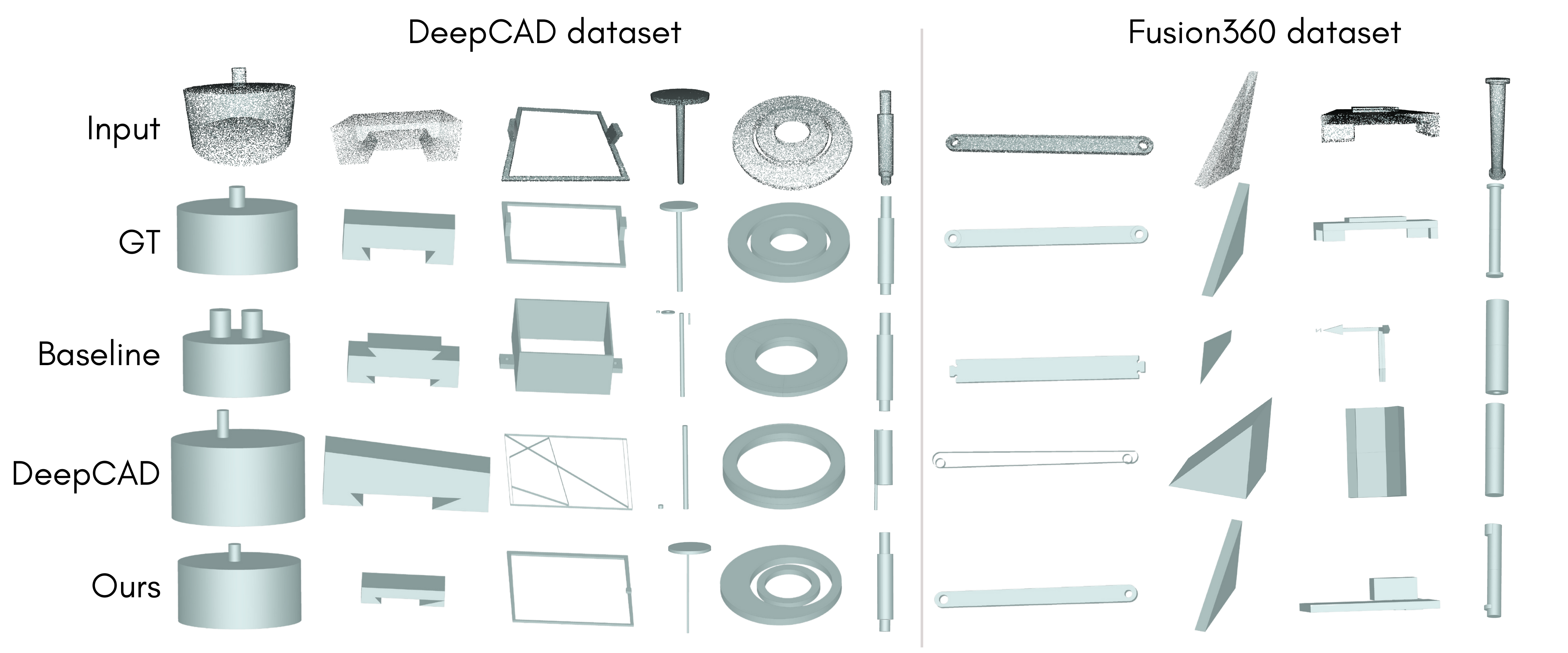}
\end{center}
 \caption{Qualitative results on the DeepCAD~\cite{wu2021deepcad} dataset (left) and the cross-dataset Fusion360~\cite{willis2021fusion} experiment (right).}
\label{fig:qualitative_results}
\end{figure}

\noindent \textbf{Qualitative Results:} Fig.~\ref{fig:qualitative_results} shows some qualitative results for the retrieval baseline, DeepCAD~\cite{wu2021deepcad} and \modelname~(Ours) on both the DeepCAD~\cite{wu2021deepcad} and Fusion360~\cite{willis2021fusion} datasets. As mentioned in Section~\ref{sec:exp_setup}, the DeepCAD dataset contains many duplicates, not just in terms of CAD sequence but also in terms of geometrical shape. As a result, the retrieval baseline is able to identify accurately duplicates (most right column of the DeepCAD dataset panel) and in other cases the baseline manages to retrieve shapes with similar geometry as the ground truth CAD model. On the other hand, it can be noticed that DeepCAD~\cite{wu2021deepcad} can even fail at retrieving duplicates. \modelname~is able to predict models that are similar in shape and also in terms of loop-extrusion sequence, even though it sometimes fails to predict the parameters accurately (second and fifth columns of DeepCAD dataset and fourth column of Fusion360 dataset).

\begingroup
\setlength{\tabcolsep}{3pt}
\begin{table}[t!]
\centering
\setlength{\belowcaptionskip}{-0.6cm}
\resizebox{\linewidth}{!}{
\begin{tabular}{ccccc}
\hline \\[-1em]
  & Model & APCS$\uparrow$ & $CD\downarrow$ & IR$\downarrow$ \\[-1em] \\  
 \hline \\[-1em]
\multirow{4}{*}{\rotatebox{90}{DeepCAD}} & \textit{Retrieval}               &   0.629               &        2.8           & 0                \\
                           & \textit{MultiCAD}~\cite{multicad}                 & -                   & 8.1               & 0.115             \\
                           & \textit{DeepCAD}~\cite{wu2021deepcad}                & 0.604                       & 19.2           & 0.038         \\
                           
                           & \textit{Ours}           & 0.732                     & 4.5            & 0.011    
                           \\[-1em]\\ \hline 

\end{tabular}
\quad       
\begin{tabular}{ccccc}
\hline \\[-1em]
  & Model & APCS$\uparrow$ & $CD\downarrow$ & IR$\downarrow$ \\[-1em] \\  
 \hline \\[-1em]
\multirow{4}{*}{\rotatebox{90}{Fusion360}}  & \textit{Retrieval}                 &  0.304                            &    60.0             & 0             \\
                           & \textit{MultiCAD}~\cite{multicad}                & -                    & 42.2             & 0.165          \\
                           
                           & \textit{DeepCAD}~\cite{wu2021deepcad}             & 0.360   & 104.2            &    0.017      \\
                           & \textit{Ours}                   & 0.365         & 33.3   & 0.024  \\[-1em]
                           \\ \hline
\end{tabular}
}
\caption{Quantitative results on the DeepCAD~\cite{wu2021deepcad} dataset and cross-dataset experiment on Fusion360~\cite{willis2021fusion}. The APCS results show that \modelname~(\textit{Ours}) is able to recover CAD sequence most accurately.}
\label{table:results}
\end{table}
\endgroup

\begin{figure}[ht]
     \centering
     \begin{subfigure}[b]{0.43\linewidth}

         \includegraphics[width=\textwidth]{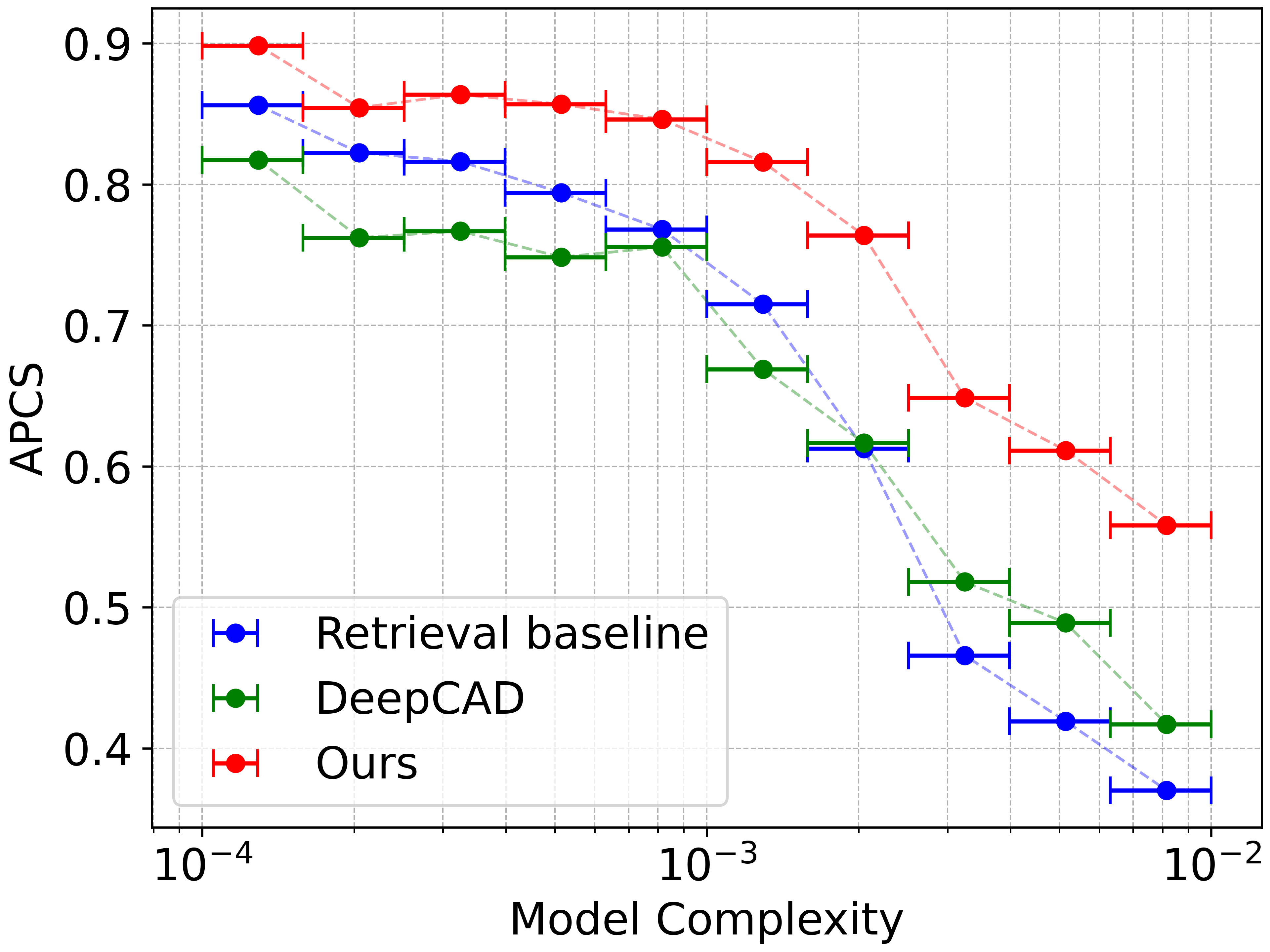}

    \end{subfigure}
    \begin{subfigure}[b]{0.43\linewidth}

         \includegraphics[width=\textwidth]{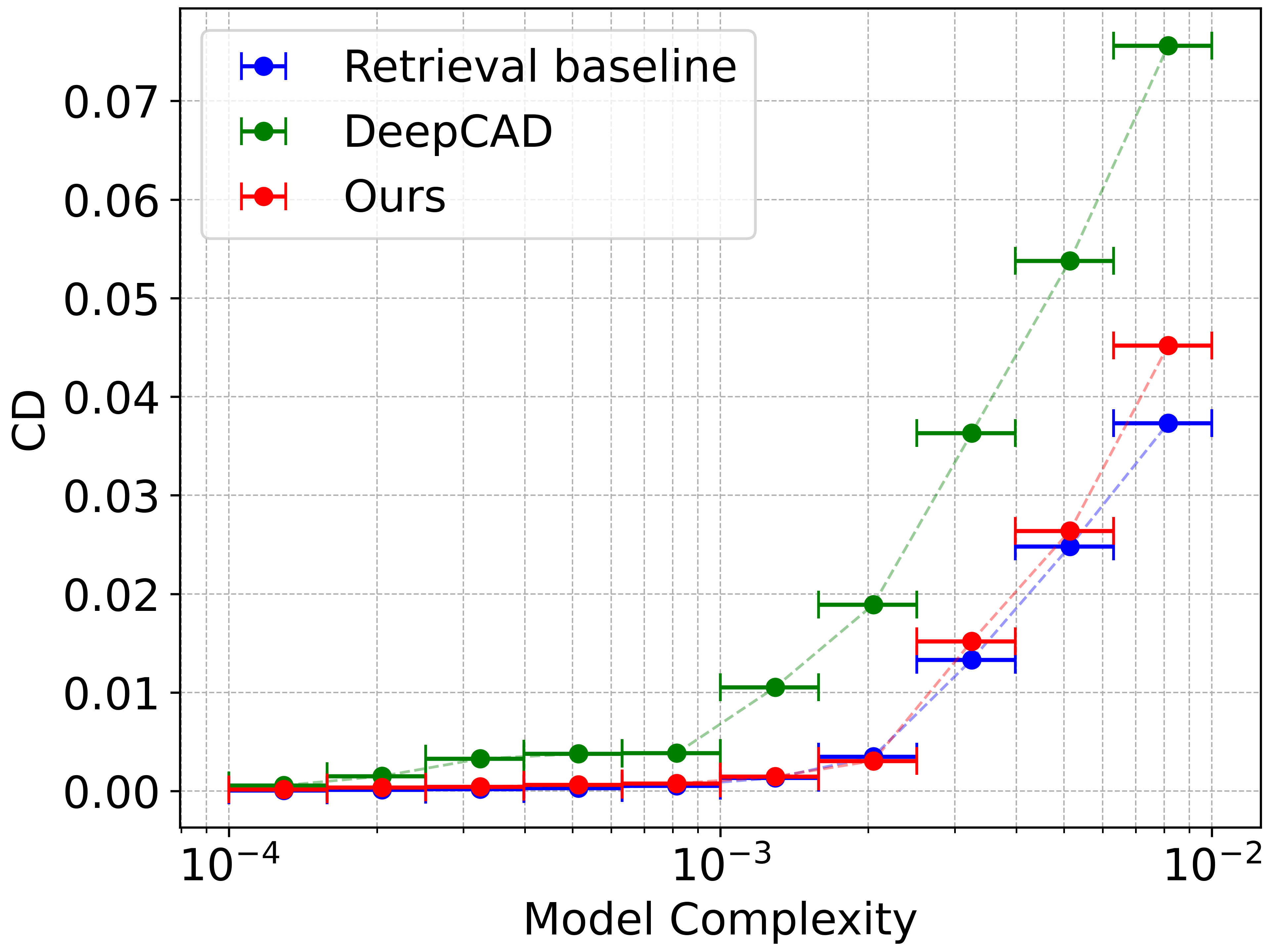}

     \end{subfigure}
        \caption{Left: Plot of the variation of the mean APCS \wrt model complexity. Right: Plot of the variation of the median $CD$ \wrt model complexity. For both graphs the number of models in each bin, represented by the horizontal bounded lines, correspond to approximately the same number of test samples from the DeepCAD dataset~\cite{wu2021deepcad}.}
        \label{fig:graphs_complex}
\end{figure}

\vspace{0.2cm}
\noindent 
\textbf{Quantitative Results:} The trends observed from the qualitative results are further supported by the quantitative results presented in Table~\ref{table:results}. The APCS, on both DeepCAD~\cite{wu2021deepcad} and Fusion360~\cite{willis2021fusion} datasets, show that \modelname~is the most capable model at predicting correct CAD sequences. However, it can be noted that the retrieval baseline obtains the lowest $CD$ by a small margin on the DeepCAD dataset and by a more significant margin on the Fusion360. One of the reasons is that this baseline always outputs a CAD model that is of roughly similar shape as the input even if the retrieved CAD sequence can vary from the ground truth. To further analyse the results, the variations of the APCS and $CD$ \wrt model complexity on the DeepCAD dataset are displayed in Fig.~\ref{fig:graphs_complex}. We define the model complexity as the lowest possible $CD$ of a test point cloud sample with respect to the train samples. In other words, the model complexity quantifies the amount by which a test sample is out of distribution from the train set in terms of shape. While \modelname~consistently outperforms on average all other baselines in terms of APCS for all model complexities, DeepCAD~\cite{wu2021deepcad} can only perform better than the retrieval baseline for the more complex models. This shows that DeepCAD~\cite{wu2021deepcad} often fails at retrieving the CAD sequence for simple models. In terms of $CD$, it can be observed that \modelname~and the retrieval baseline have similar performance. \modelname~can predict a CAD sequence that is closer to the ground truth one but the predicted overall shape can vary from the ground truth for more difficult samples.

\vspace{0.2cm}
\noindent 
\textbf{Ablation Study:} In this section, the different components of the proposed network architecture are ablated. Table~\ref{table:ablation_main} shows the results for \textit{Ours w/o hier.}, in which the learning is done without both the loop-extrusion decoder and the refining network, \textit{Ours w/o refining} where the refining component is ablated and \textit{Ours}. It can be noted that each component leads to an improvement in all the metrics. It is worth noting that the $F1$ score on the loop-extrusion type prediction introduced for the hierarchical learning of \modelname~is $0.79$. This implies that on most cases both loop and extrusion decoders receive embedding of the correct type. Moreover, while the refining network is only applied on the loop parameters, we observe that the component of the APCS for the extrusion parameters are also higher when the network is trained with the refining component. This suggests that the refining network is able to provide a useful signal for guiding the learning process.  More details are in the supplementary material.

\begingroup
\setlength{\tabcolsep}{3pt}
\begin{table}[t!]
\centering
\setlength{\belowcaptionskip}{-0.4cm}
\begin{tabular}{cccc}
\hline \\[-1em]
   Model & APCS$\uparrow$ &  $CD\downarrow$ & IR$\downarrow$ \\[-1em] \\  
 \hline \\[-1em]
                            \textit{Ours w/o hier.}                &       0.687         &               7.0 &    0.016         \\
                            \textit{Ours w/o refining}               &       0.708         &       4.8   &   0.018      \\
                           
                            \textit{Ours}           &   0.732              &  4.5           &     0.011
                           \\[-1em]\\ \hline \\[-1em]
                           
\end{tabular}
\caption{Ablation results demonstrating the relevance of the hierarchical learning strategy and the loop refiner.}
\label{table:ablation_main}
\end{table}
\endgroup

\vspace{0.2cm}
\noindent
\textbf{Input point cloud perturbation}: 
Reverse engineering is a real-world practical problem. The results in the previous section are obtained from sampling points from the B-Rep representations of the CAD models. While modern 3D sensors can reconstruct the mesh of models with high resolution, they still suffer from some artifacts such as noise and small missing parts. In order to evaluate the performance of our network in such realistic conditions, we run experiments in two scenarios, one in which noise is added and one in which small holes are created on the point cloud. In order to simulate realistic noise, Perlin noise~\cite{perlin_noise} is added to the mesh from which the point coordinates and normals are extracted. More details about the noise and hole generations can be found in the supplementary materials. Table~\ref{table:pertubation_results} shows that \modelname~is more robust to such perturbations than other methods. As the noise also adds a disturbance to the direction of the input point normals, it leads to a larger drop in performance.

\begingroup
\setlength{\tabcolsep}{3pt}
\begin{table}[t!]
\centering
\setlength{\belowcaptionskip}{-0.5cm}
\resizebox{0.9\linewidth}{!}{
\begin{tabular}{ccccc}
\hline \\[-1em]
  & Model & APCS$\uparrow$ & $CD\downarrow$ & IR$\downarrow$ \\[-1em] \\  
 \hline \\[-1em]
\multirow{3}{*}{\rotatebox{90}{Noise}} & \textit{Retrieval}               &       0.561           &               5.9                &     0       \\
                          
                           & \textit{DeepCAD}~\cite{wu2021deepcad}                &      0.550      &         31.1      &     0.047     \\
                           
                           & \textit{Ours}           &       0.604     &     18.1        &     0.010
                           \\[-1em]
                           \\ \hline 
\end{tabular}
\quad    
\begin{tabular}{ccccc}
\hline \\[-1em]
  & Model & APCS$\uparrow$ & $CD\downarrow$ & IR$\downarrow$ \\[-1em] \\  
 \hline \\[-1em]
\multirow{3}{*}{\rotatebox{90}{Holes}}  & \textit{Retrieval}                 &     0.607           &               4.1         & 0             \\
                          
                           & \textit{DeepCAD}~\cite{wu2021deepcad}             &  0.574   &   26.7         &      0.039    \\
                           & \textit{Ours}                   &    0.732    &  4.4  & 0.012  \\[-1em]
                           \\ \hline
\end{tabular}
}
\caption{Results on the DeepCAD~\cite{wu2021deepcad} dataset when the input point cloud is perturbated either with Perlin noise (left) or by creating holes (right).}
\label{table:pertubation_results}
\end{table}
\endgroup

\begin{figure}[ht!]
\setlength{\belowcaptionskip}{-0.5cm}
\begin{center}
\includegraphics[width=\linewidth]{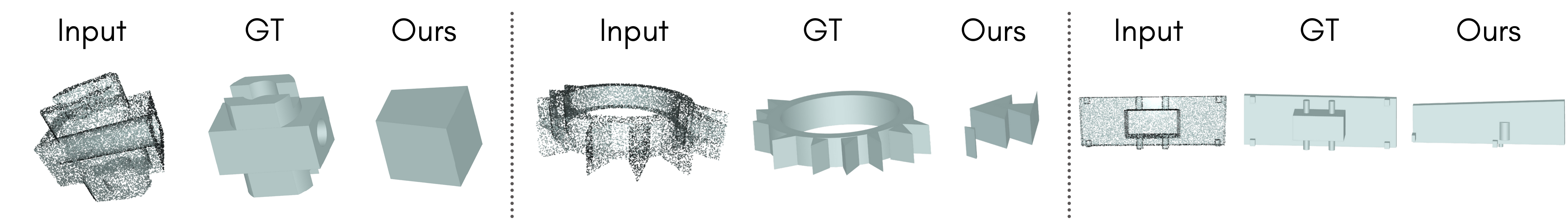}
\end{center}

 \caption{Examples for which \modelname~fails at recovering a shape close to the ground truth one.}
\label{fig:failure}
\end{figure}

\vspace{0.2cm}
\noindent
\textbf{Failure Cases and Limitations}: In this section, we describe the reasons that lead \modelname~to predict invalid CAD models as measured by IR. Among the predictions of \modelname~ on DeepCAD test set, only one contains a loop parametrization that results in an invalid CAD sequence. This shows that the proposed representation of the loop sequence leads to syntactically correct loops on practically all cases. However, in some cases the representation of the CAD sequence can be syntactically valid, yet it is not possible to reconstruct a B-Rep from it. For 75 test samples, the loop-extrusion decoder fails to predict an extrusion token, this implies that the predicted model is therefore an infinitely thin sketch and not a 3D model as expected. The rest of the invalid models are mostly due to a loop being made of a single line within a model, which cannot be extruded into a valid 3D shape within our context. Finally, Fig.~\ref{fig:failure} shows examples for which the predicted sequences do not lead to a shape that is close to the ground truth one. In these examples, the input CAD models contain a large number of small features that \modelname~is unable to capture.

\section{Conclusion}
\label{sec:conclusion}

In conclusion, we propose \modelname~an end-to-end transformer-based neural network that learns to recover the CAD sequence from a given point cloud. Two of the main features of \modelname~are a hierarchical structure that enables the learning of a high-level loop-extrusion sequence and a loop refiner that aims at correcting errors in loop parameter predictions. We also propose a primitive representation in which each primitive is described by the same number of parameters. We identify the limitations of current metrics in the emerging domain of 3D reverse engineering and propose a new metric, the APCS that leads to a fair comparison of parametric CAD sequences. Thorough experiments show that \modelname~achieves state-of-the-art results in different realistic scenarios.

\vspace{0.2cm}
\noindent \textbf{Acknowledgement:} The present project is supported by the National Research Fund, Luxembourg under the BRIDGES2021/IS/16849599/FREE-3D and \\IF/17052459/CASCADES projects, and by Artec 3D.

%
%
\bibliographystyle{splncs04}
\bibliography{egbib}

\clearpage
\setcounter{section}{0}
\setcounter{figure}{0}
\setcounter{table}{0}
\renewcommand{\thesection}{\Alph{section}}

\midtitle{Supplementary Materials\\ \modelname: A Hierarchical Transformer for CAD Sequence Inference from Point Clouds}

\section{Formulation}

\begin{figure}[ht]
\begin{center}
\includegraphics[width=\linewidth]{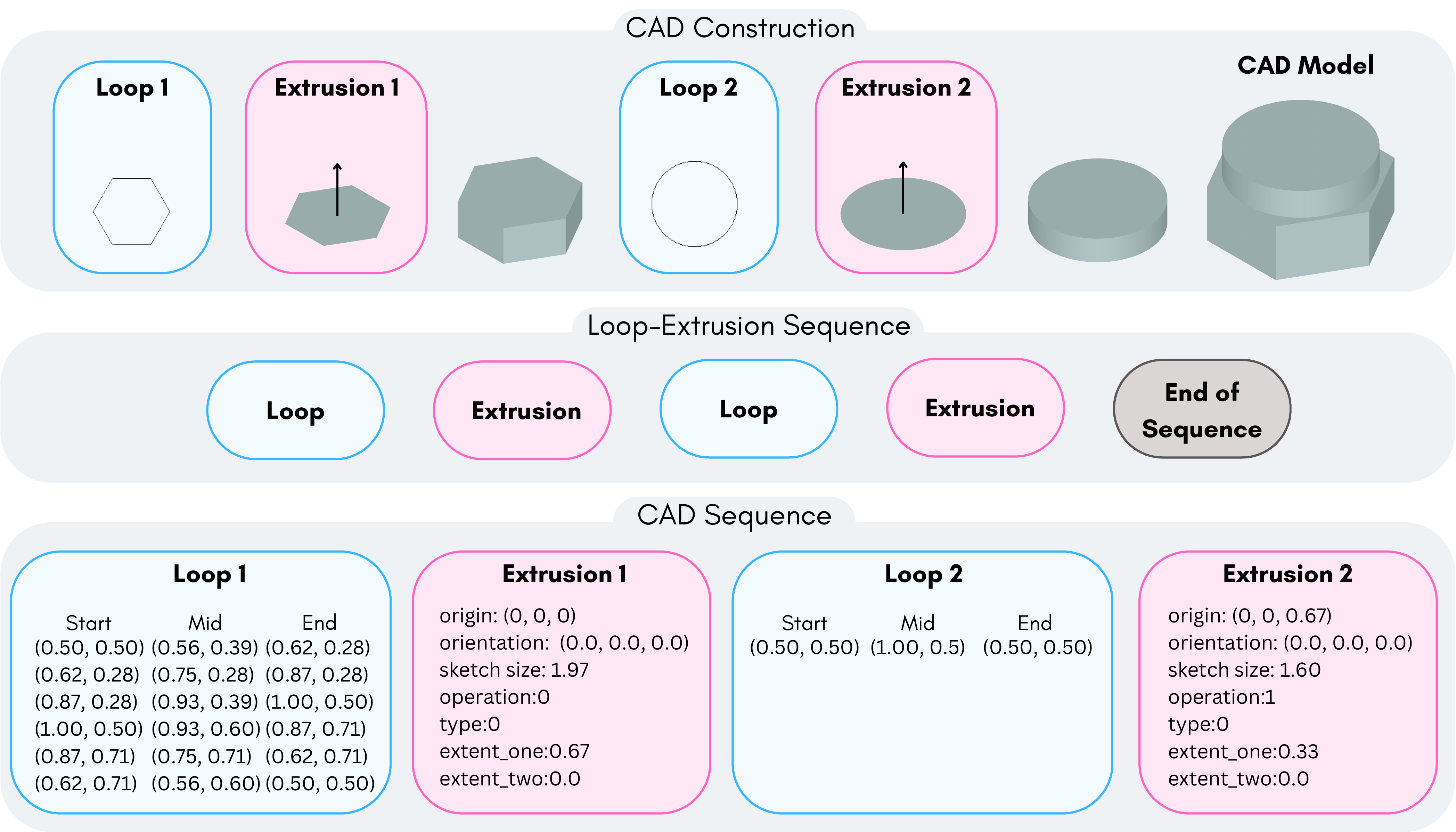}
\end{center}
 \caption{Example of a CAD model and its CAD sequence in the formulation proposed. The top panels depicts the CAD construction process. The middle panel shows the corresponding high-level loop-extrusion sequence that is predicted as part of the hierarchical learning. The low-level parameters of each loop primitive and extrusion are displayed in the bottom panel.}
\label{fig:formulation}
\end{figure}

Fig.~\ref{fig:formulation} shows an example of the construction process of a CAD model as well as the corresponding loop-extrusion sequences. The loop-extrusion sequence is made of three types of tokens (loop, extrusion and end of sequence). The CAD sequence combines the high level loop-extrusion sequence tokens and their corresponding parameters. Note for the loop parameters, the coordinates of the start point of a primitive is always the same as the coordinates of the end point of the previous primitive as in~\cite{wu2021deepcad}.

\section{Duplicates in the DeepCAD Dataset}

\begin{figure}[ht]
\begin{center}
\includegraphics[width=\linewidth]{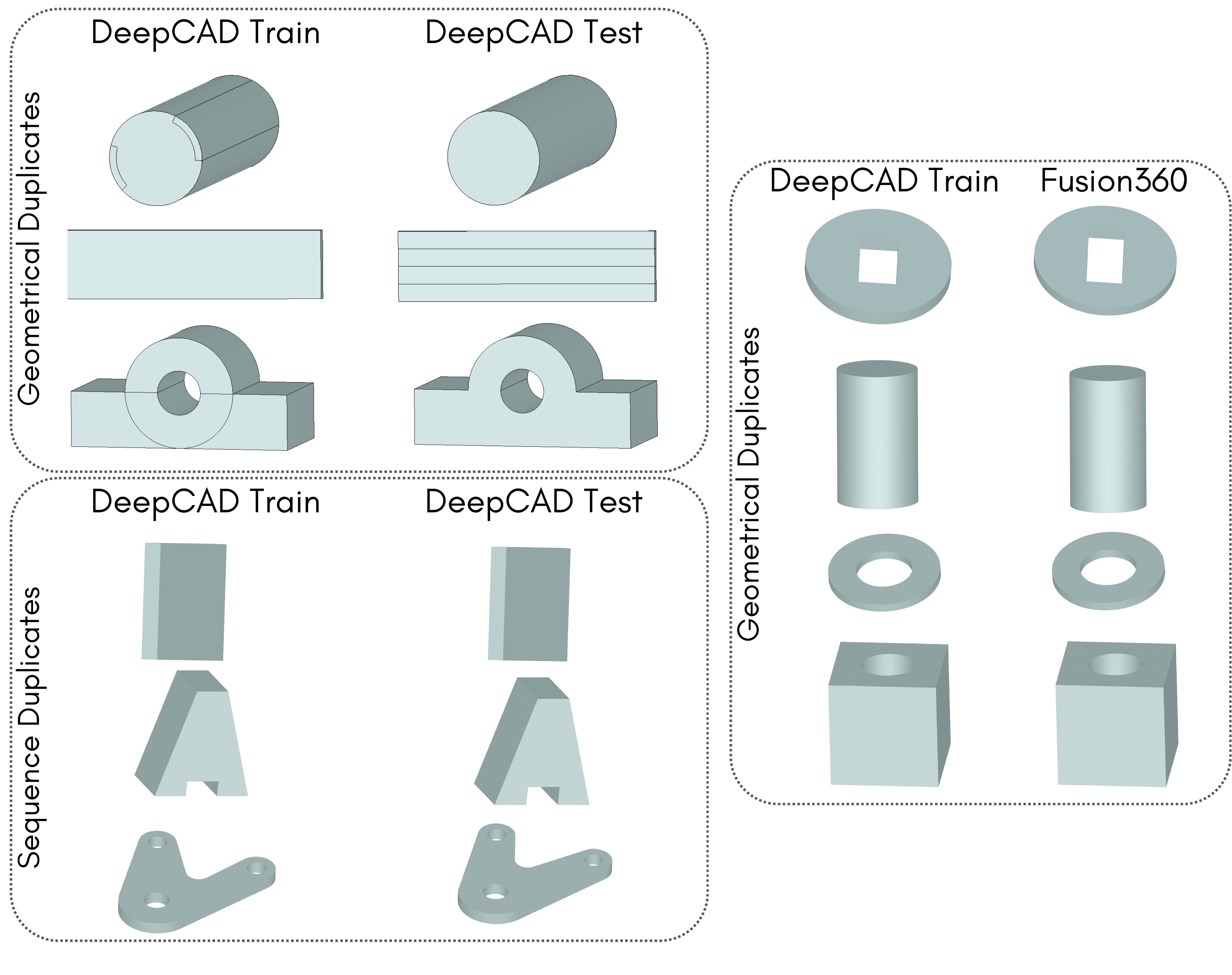}
\end{center}
 \caption{Examples of duplicate CAD models from the DeepCAD~\cite{wu2021deepcad} and Fusion360~\cite{willis2021fusion} datasets. On the top left panel, CAD models from the DeepCAD train set with geometrical duplicates in the test set are shown. Similarly, the right panel presents geometrical duplicates present in the Fusion360~\cite{willis2021fusion} dataset. CAD models with identical CAD sequences, \ie sequence duplicates, are displayed in the bottom left panel.}
\label{fig:duplicates}
\end{figure}

As mentioned in Section~\ref{sec:exp_setup} of  the main paper, one of the main limitations of the DeepCAD dataset~\cite{wu2021deepcad} is that it contains many duplicates across the train and test sets. While the works in~\cite{xu2022skexgen} proposed a strategy to identify duplicates that have exactly the same CAD sequence (\ie \textit{sequence duplicates}), we observe that the dataset also contains models that have almost identical geometry but different CAD sequences (\ie \textit{geometrical duplicates}). Moreover some CAD models in the DeepCAD train set are almost identical to some models in the Fusion360 dataset~\cite{willis2021fusion}, with often just the amount of extrusion varying slightly. Fig.~\ref{fig:duplicates} shows examples of the different types of duplicates from the DeepCAD~\cite{wu2021deepcad} and Fusion360~\cite{willis2021fusion} datasets.
We define a geometrical duplicate as a test set CAD model for which it exists a CAD model in the train set with a chamfer distance less than the uncertainty in the chamfer distance measurement ($\pm 3\times10^{-4}$ when $4096$ points are sampled). From this definition and using the train set of the DeepCAD dataset~\cite{wu2021deepcad}, we observe that about $14\%$ of the DeepCAD test set is made of geometrical duplicates and about $12\%$ in the Fusion360 test set. We observe that current datasets that contain both CAD models and their corresponding CAD sequences are limited by either by the number of samples (Fusion360~\cite{willis2021fusion}) or by the lack of diversity they present (DeepCAD~\cite{wu2021deepcad}).

\section{Implementation details}
In this section, more details on the network architecture are provided.

\vspace{0.2cm}
\noindent \textbf{Transformer decoders:} The loop-extrusion decoder $\mathbf{\Phi}_{\rho,e}$ and the loop decoder $\mathbf{\Phi}_{\rho}$ are both transformer decoder with the same network architecture. They are both made of $4$ layers, each made of $8$ heads with a feed-forward dimension of $512$. A dropout rate of $0.1$ is used.

\vspace{0.2cm}
\noindent \textbf{PointNet++:} As mentioned in the main paper, the point encoder  $\mathbf{\Phi}_p$ is PointNet++~\cite{qi2017pointnet++}. The implementation provided in~\cite{pytorchpointnet++} was used. The input dimension $d_p=6$ corresponds to the point and normal coordinates. The parameters for the $4$ layers are as follows: number of points ($512$, $256$, $128$, $16$) with radius ($0.1$, $0.2$, $0.4$, $0.8$) and number of samples ($64$, $64$, $64$, $32$).

\section{APCS Metric}
In this section, more details on the computation of the proposed metric APCS (see Section~\ref{sec:evaluation_framework} of the main paper) are provided.

The \textit{mean Average Precision of CAD Sequence}, APCS, is a score between $0$ and $1$ that evaluates how close two CAD sequences are by taking into account the types and parameters of loop primitives and the extrusion parameters. The types of primitives considered in this work are \textit{line}, \textit{arc} and \textit{circle}. The parameters of the loop primitives correspond to $6$ point coordinates in a normalized 2D space. An extrusion is described using $11$ parameters as in~\cite{wu2021deepcad} which can be grouped into $4$ main categories: 1) Ext.: the type of extrusion and the amount of extrusion; 2) Origin: the origin of the 2D sketch in 3D space; 3) Orientation: the sketch plane orientation; and 4) Size: the size of the sketch in 3D space. Note the latter $3$ categories describe the projection and scaling of the normalized 2D sketch in 3D.

\section{Results}
In this section, further quantitative and qualitative results are presented.

\subsection{Quantitative Results}

The APCS scores reported in the main paper are computed per model and then averaged over all the models of the test set. It is also possible to compute the average of each individual component of the APCS scores over the whole test set. Such results are discussed in the following paragraphs for the DeepCAD dataset~\cite{wu2021deepcad}, Fusion360 dataset~\cite{willis2021fusion} and the ablation study.

\noindent \textbf{DeepCAD Dataset:} Table~\ref{table:Deepcad} shows the APCS scores on the DeepCAD dataset~\cite{wu2021deepcad} for each of the components averaged over the test set. It can be observed that \modelname~(\textit{Ours}) obtains a significantly better score for the arc primitive and also to some extent for the circle primitive compared to the retrieval baseline and DeepCAD~\cite{wu2021deepcad}. However, the scores corresponding to the placement of the 2D sketch in 3D (Origin, Orientation and Size) for \modelname~are slightly lower than for DeepCAD~\cite{wu2021deepcad}.

\begingroup
\setlength{\tabcolsep}{3pt}
\begin{table}[t!]
\centering
\begin{tabular}{cccccccc}
\hline \\[-1em]
\multicolumn{1}{c}{\multirow{2}{*}{Model}} & \multicolumn{7}{c}{APCS$\uparrow$ }\\
   & Line &  Arc& Circle&  Ext.& Origin & Orientation& Size \\[-1em] \\  
 \hline \\[-1em]
                     \textit{Retrieval}               &       0.584        &       0.280 &   0.655   &0.716&0.666&0.819&0.691     \\
                               \textit{DeepCAD}~\cite{wu2021deepcad}               &       0.654       &      0.246   &   0.587  &0.872&0.825&0.928&0.848    \\
                           
                            \textit{Ours}           &   0.665              &  0.709         & 0.683&0.818&0.768&0.879&0.778
                           \\[-1em]\\ \hline \\[-1em]
                           
\end{tabular}
\caption{Results of the different APCS components on the DeepCAD dataset~\cite{wu2021deepcad}.}
\label{table:Deepcad}
\end{table}
\endgroup

The APCS score provides a more complete evaluation of the predicted CAD sequences than the command accuracy ($ACC_{cmd}$) and parameter accuracy ($ACC_{param}$) used in the works~\cite{wu2021deepcad}. Nevertheless, we provide the results for \modelname~and DeepCAD~\cite{wu2021deepcad} against those metrics in Table~\ref{table:Deepcad_metric} for the DeepCAD dataset.

\begingroup
\setlength{\tabcolsep}{3pt}
\begin{table}[t!]
\centering
\begin{tabular}{ccc}
\hline \\[-1em]
Model & $ACC_{cmd}$$\uparrow$ 
   & $ACC_{param}$$\uparrow$  \\  
 \hline \\[-1em]

                               \textit{DeepCAD}~\cite{wu2021deepcad}               &      0.821       &      0.693       \\
                           
                            \textit{Ours}           &   0.838            &  0.741       
                           \\[-1em]\\ \hline \\[-1em]
                           
\end{tabular}
\caption{Results using the metrics described in DeepCAD on the DeepCAD dataset~\cite{wu2021deepcad}.}
\label{table:Deepcad_metric}
\end{table}
\endgroup

\begin{figure}[ht]
     \centering
     \begin{subfigure}[b]{0.47\linewidth}

         \includegraphics[width=\textwidth]{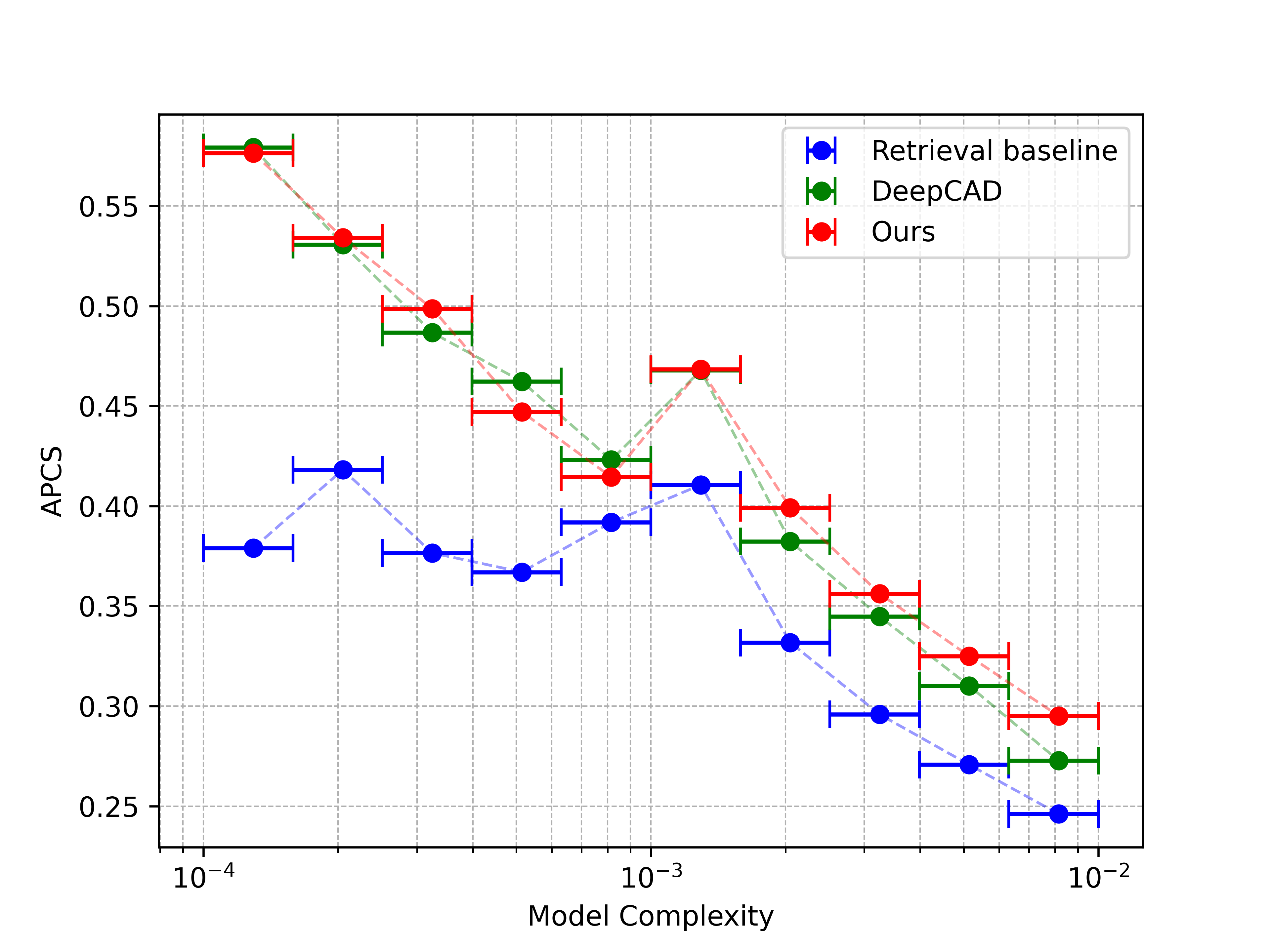}

    \end{subfigure}
    \begin{subfigure}[b]{0.47\linewidth}
         \includegraphics[width=\textwidth]{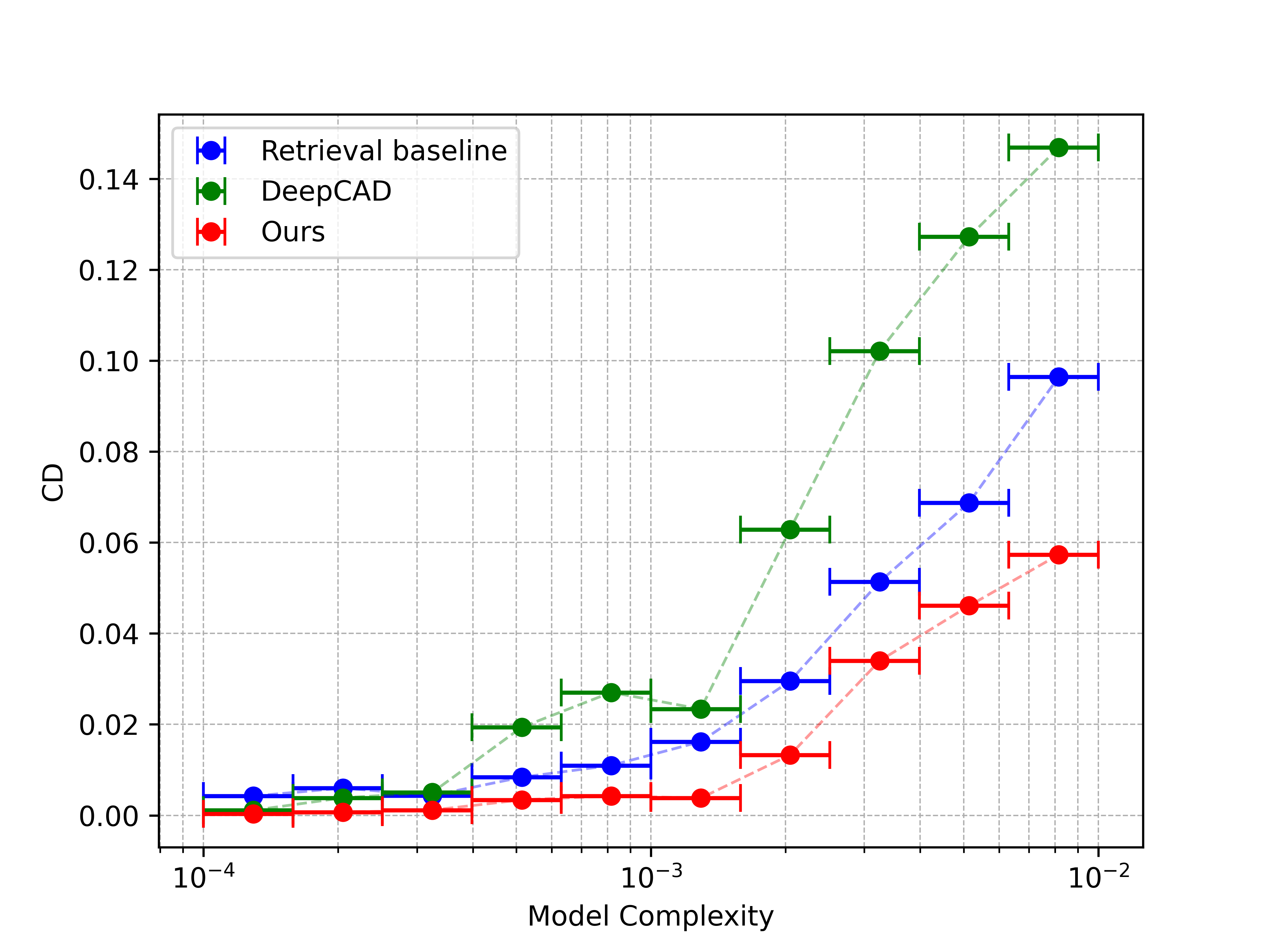}

     \end{subfigure}
        \caption{Left: Plot of the variation of the mean APCS \wrt model complexity. Right: Plot of the variation of the median $CD$ \wrt model complexity. For both graphs the number of models in each bin, represented by the horizontal bounded lines, correspond to approximately the same number of test samples from the Fusion360 dataset~\cite{willis2021fusion}.}
        \label{fig:graphs}
\end{figure}

\vspace{0.2cm}
\noindent \textbf{Fusion360 Dataset:} The trends previously described on the DeepCAD dataset~\cite{wu2021deepcad} can also be observed on the cross-dataset evaluation using the Fusion360 dataset~\cite{willis2021fusion} (see Table~\ref{table:Fusion}). Furthermore, the APCS score corresponding to the line primitive is significantly higher for \modelname~than for the other two baseline models. Fig.~\ref{fig:graphs} shows the variation of the APCS and \textit{CD} \wrt to model complexity for the Fusion360 dataset~\cite{willis2021fusion}. It can be observed that while the performance on the APCS metric for DeepCAD~\cite{wu2021deepcad} and \modelname~are relatively close, \modelname~achieves lowest $CD$ for all model complexities compared to DeepCAD~\cite{wu2021deepcad} and the retrieval baseline.

\begingroup
\setlength{\tabcolsep}{3pt}
\begin{table}[t!]
\centering
\begin{tabular}{cccccccc}
\hline \\[-1em]
\multicolumn{1}{c}{\multirow{2}{*}{Model}} & \multicolumn{7}{c}{APCS$\uparrow$ }\\
   & Line &  Arc& Circle&  Ext.& Origin & Orientation& Size \\[-1em] \\  
 \hline \\[-1em]
                     \textit{Retrieval}               &       0.214        &               0.045 &   0.546 &0.453&0.294&0.509&0.451     \\
                               \textit{DeepCAD}~\cite{wu2021deepcad}               &      0.368       &      0.144  &  0.639  &0.844&0.774&0.866&0.859    \\
                           
                            \textit{Ours}           &    0.515             &      0.383  & 0.622&0.749&0.693&0.836&0.782
                           \\[-1em]\\ \hline \\[-1em]
                           
\end{tabular}
\caption{Results of the different APCS components on the Fusion dataset~\cite{willis2021fusion}.}
\label{table:Fusion}
\end{table}
\endgroup

\begin{figure}[ht]
\begin{center}
\includegraphics[width=0.8\linewidth]{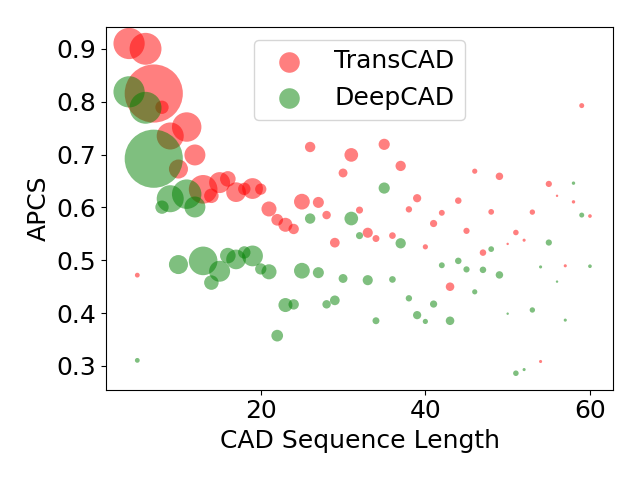}
\end{center}
 \caption{APCS results as a function of the ground truth CAD sequence length expressed in the DeepCAD format on the DeepCAD dataset~\cite{wu2021deepcad}. The size of the points is proportional to the number of CAD models with the corresponding CAD sequence length.}
\label{fig:apcs_vs_cadseq}
\end{figure}

\begin{figure}[ht]
\begin{center}
\includegraphics[width=0.8\linewidth]{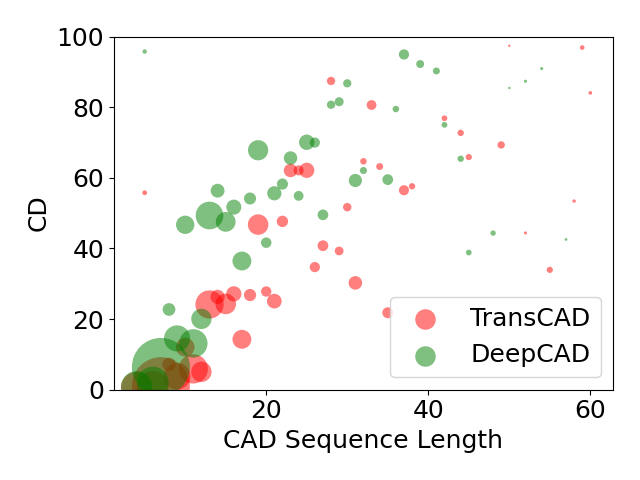}
\end{center}
 \caption{\textit{CD} results as a function of the ground truth CAD sequence length expressed in the DeepCAD format on the DeepCAD dataset~\cite{wu2021deepcad}. The size of the points is proportional to the number of CAD models with the corresponding CAD sequence length.}
\label{fig:cd_vs_cadseq}
\end{figure}

\noindent \textbf{Complex shape performance:} 
Figure~\ref{fig:apcs_vs_cadseq} show the APCS \wrt the sequence length for \modelname~and DeepCAD~\cite{wu2021deepcad}. Similarly, Figure~\ref{fig:cd_vs_cadseq} presents the variation of $CD$ \wrt the sequence length. The size of the data points is proportional to the number of models it represents. The performance decreases for both models as the length of the CAD sequence increases. However, \modelname~consistently outperforms DeepCAD.

\vspace{0.2cm}
\noindent \textbf{Ablation Study:} Table~\ref{table:ablation} shows the APCS component scores corresponding to the ablation study presented in Section~\ref{sec:exp_results} of the main paper. The most striking result is that the Loop Refiner leads to improved performance in all the extrusion components, even though the Loop Refiner only acts directly on the loop parameters. The Loop Refiner is a component of the end to end pipeline of \modelname. As a result, the backprogation of the Loop Refiner loss $\mathcal{L}_{\rho}$ acts on all the parameters of the network and can therefore impact the predictions at all levels.

\begingroup
\setlength{\tabcolsep}{3pt}
\begin{table}[t!]
\centering
\begin{tabular}{cccccccc}
\hline \\[-1em]
\multicolumn{1}{c}{\multirow{2}{*}{Model}} & \multicolumn{7}{c}{APCS$\uparrow$ }\\
    & Line &  Arc& Circle&  Ext.& Origin & Orientation& Size \\[-1em] \\  
 \hline \\[-1em]
                            \textit{Ours w/o hier.}                &       0.569        &             0.473 &   0.529   &0.749&0.640&0.843&0.670     \\
                            \textit{Ours w/o refining}               &        0.655        &   0.625      &  0.673  &0.755&0.679&0.838&0.703    \\
                           
                            \textit{Ours}           &            0.665  &   0.709       & 0.683&0.818&0.768&0.879&0.778
                           \\[-1em]\\ \hline \\[-1em]
                           
\end{tabular}
\caption{Results of the different APCS components on the DeepCAD dataset~\cite{wu2021deepcad} for the ablation study. \textit{Ours w/o hier.} corresponds the proposed model without hierarchical learning, and Loop Refiner and \textit{Ours w/o refining} to the proposed model without the Loop Refiner.}
\label{table:ablation}
\end{table}
\endgroup

 The Loop-Extrusion module classifies the features $\mathbf F_{\rho,e}$ as \textit{loop}, \textit{extrusion} or  \textit{end of sequence}  type. These predictions are used to route the features $\mathbf F_{\rho,e}$ to either the loop decoder $\mathbf \Phi_{\rho}$ or extrusion decoder $\mathbf \Phi_{e}$ to obtain their parameters. To demonstrate the impact of the Loop-Extrusion type classification, we conduct the following experiment: the ground truth Loop-Extrusion type labels are used instead of the predicted ones at testing time on the DeepCAD dataset~\cite{wu2021deepcad}. In this scenario, the APCS metric evaluating the CAD sequence increases from $0.732$ to $0.790$. The \textit{IR} also improves and decreases to nearly $0\%$ (only 2 invalid models). Notably, there is no significant change in the \textit{CD}. As a result, \modelname~is robust to moderate classification errors in the Loop-Extrusion prediction \wrt the final reconstruction. However, these errors might impact the performance of the predicted CAD sequence \wrt the ground truth. This suggests that the Loop-Extrusion classification errors might result in alternative yet plausible design paths.

\subsection{Qualitative Results}
In this section, further qualitative results are presented by first comparing \textit{TransCAD} to the retrieval baseline and then to DeepCAD~\cite{wu2021deepcad}.

\begin{figure}[ht]
\begin{center}
\includegraphics[width=0.8\linewidth]{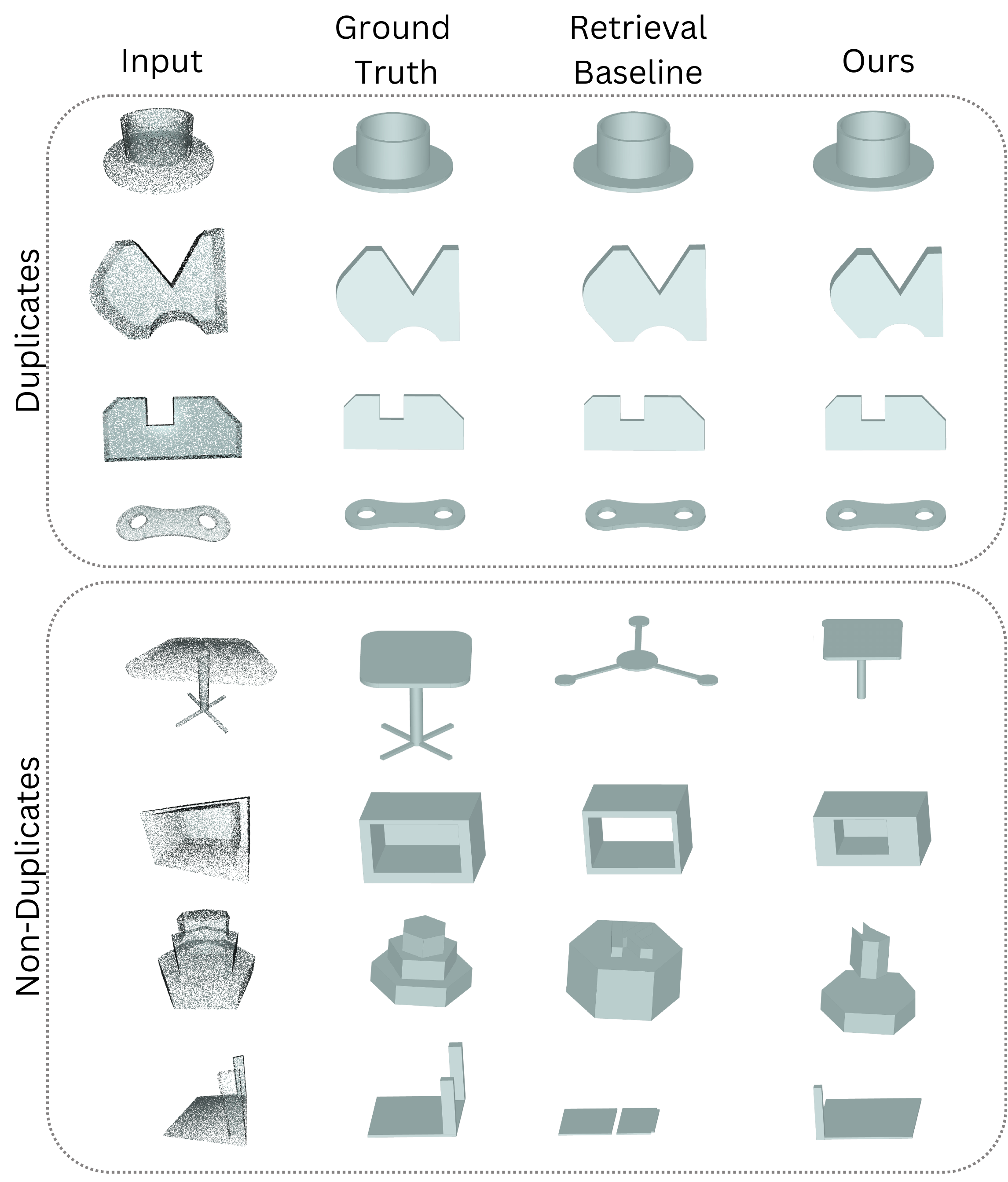}
\end{center}
 \caption{Qualitative results showing the performance of \modelname~against the retrieval baseline on both duplicate CAD models (top panel) and non-duplicate models (bottom panel).}
\label{fig:quali_baseline}
\end{figure}

\vspace{0.2cm}
\noindent \textbf{Retrieval Baseline Comparison:} Qualitative results comparing \modelname~to the the retrieval baseline can be found in Fig.~\ref{fig:quali_baseline}. It can be observed that \textit{TransCAD} can perform well on duplicate models and more importantly that it performs better than the retrieval baseline on non-duplicate models. \textit{TransCAD} can identify most of the components of an unseen CAD model but sometimes fails to place those parts in the right location as suggested by the APCS scores presented in the previous section. 

\begin{figure}[ht]
\begin{center}
\includegraphics[width=0.8\linewidth]{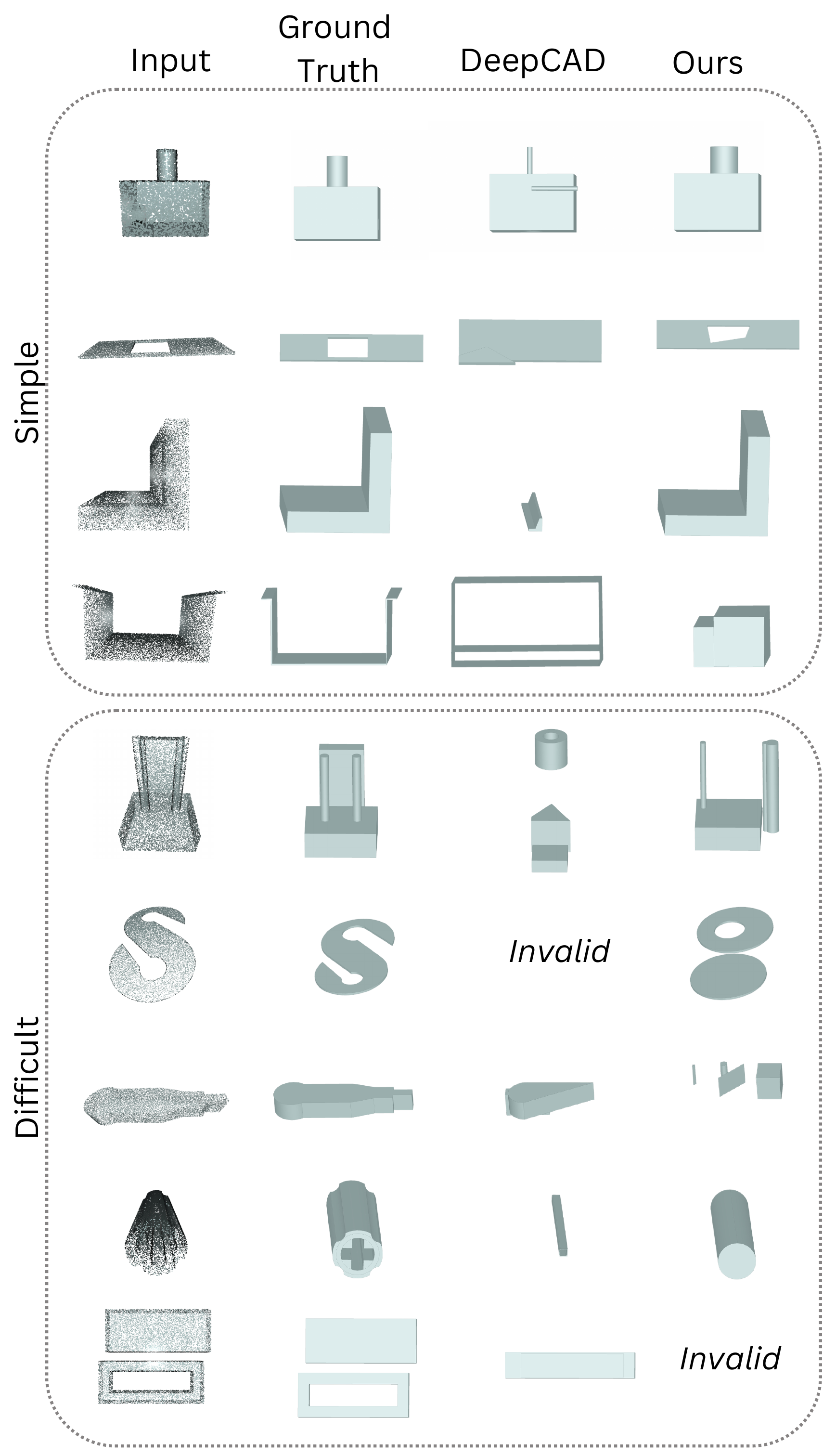}
\end{center}
 \caption{Qualitative results showing the performance of \modelname~against the DeepCAD~\cite{wu2021deepcad} on both simpler CAD models (top panel) and more complex models (bottom panel).}
\label{fig:quali_deepcad}
\end{figure}

\vspace{0.2cm}
\noindent \textbf{DeepCAD Comparison:} Qualitative results showing the comparison between \modelname~and DeepCAD~\cite{wu2021deepcad} can be found in Fig.~\ref{fig:quali_deepcad}. The results on \textit{simple} and \textit{difficult} models show that \modelname~is able to outperform DeepCAD~\cite{wu2021deepcad} on most occasions.

\section{Comparison with ComplexGen}

The output of \modelname~is a CAD sequence that allows full integration and editability in CAD software at both the final shape and intermediate design process levels. On the other hand, the output of ComplexGen~\cite{guo2022complexgen} is drastically different since it is only the final shape as a set of corners, curves, and patches with topology constraints. ComplexGen cannot predict the intermediate design steps, as shown in Fig.~1 in the paper. To the best of our knowledge, DeepCAD~\cite{wu2021deepcad} and MultiCAD~\cite{multicad} are the only works that predict CAD sequences from point clouds, with only DeepCAD providing source codes for extensive comparisons. Nevertheless, we ran a pretrained ComplexGen model on the DeepCAD test set and recorded a \textit{CD} of $2.3$ (compared to $4.5$ for \modelname). Note that as ComplexGen does not predict the CAD sequence, the APCS metric cannot be computed. On the other hand, ComplexGen metrics are tailored to their output and designed to evaluate the predicted patches of the final shape and topological validity. To conclude, other reverse engineering baselines (\eg ComplexGen~\cite{guo2022complexgen}) might achieve a better final reconstruction than \modelname, but could not be used seamlessly for reverse engineering (\ie intermediate design level editability).

\section{Point Cloud Perturbation}

In this section, further details on the implementation of the perturbations applied to the input point cloud $\textbf{P}$ described in Section~\ref{sec:exp_results} of the main paper are presented.

\subsection{Perlin Noise Implementation}

\begin{figure}[ht]
\begin{center}
\includegraphics[width=\linewidth]{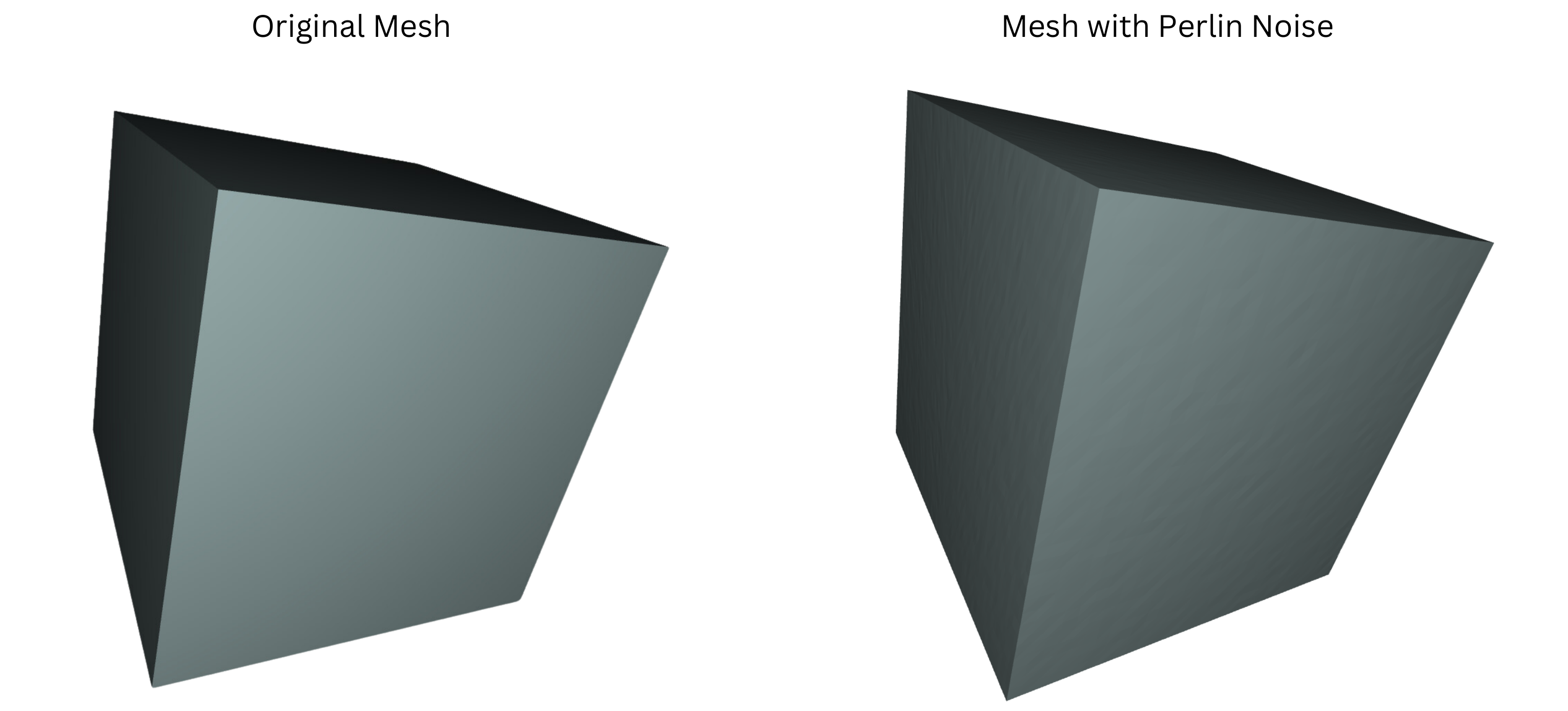}
\end{center}
 \caption{Example of a CAD model with its mesh representation (left) and perturbed representation with Perlin noise (right). Zooming on the figure might be required to best view the effect of the Perlin noise applied.}
\label{fig:perlin}
\end{figure}

To simulate the noise created when an object is scanned using a 3D sensor, a Perlin noise~\cite{perlin_noise} is added to the point cloud. The perlin noise is created using the following strategy. Starting from the mesh representation of the original CAD models, the faces are divided to ensure that the mesh contains a dense number of vertices. Then, a 3D Perlin noise is computed for each vertex using $64$ octaves with a minimum and maximum magnitude of $-0.001$ and $0.001$,  respectively. Finally, the normals are recomputed from the mesh and points are sampled. A visual example of a perturbed mesh can be found in Fig.~\ref{fig:perlin}.

\subsection{Holes Implementation}

\begin{figure}[ht]
\begin{center}
\includegraphics[width=\linewidth]{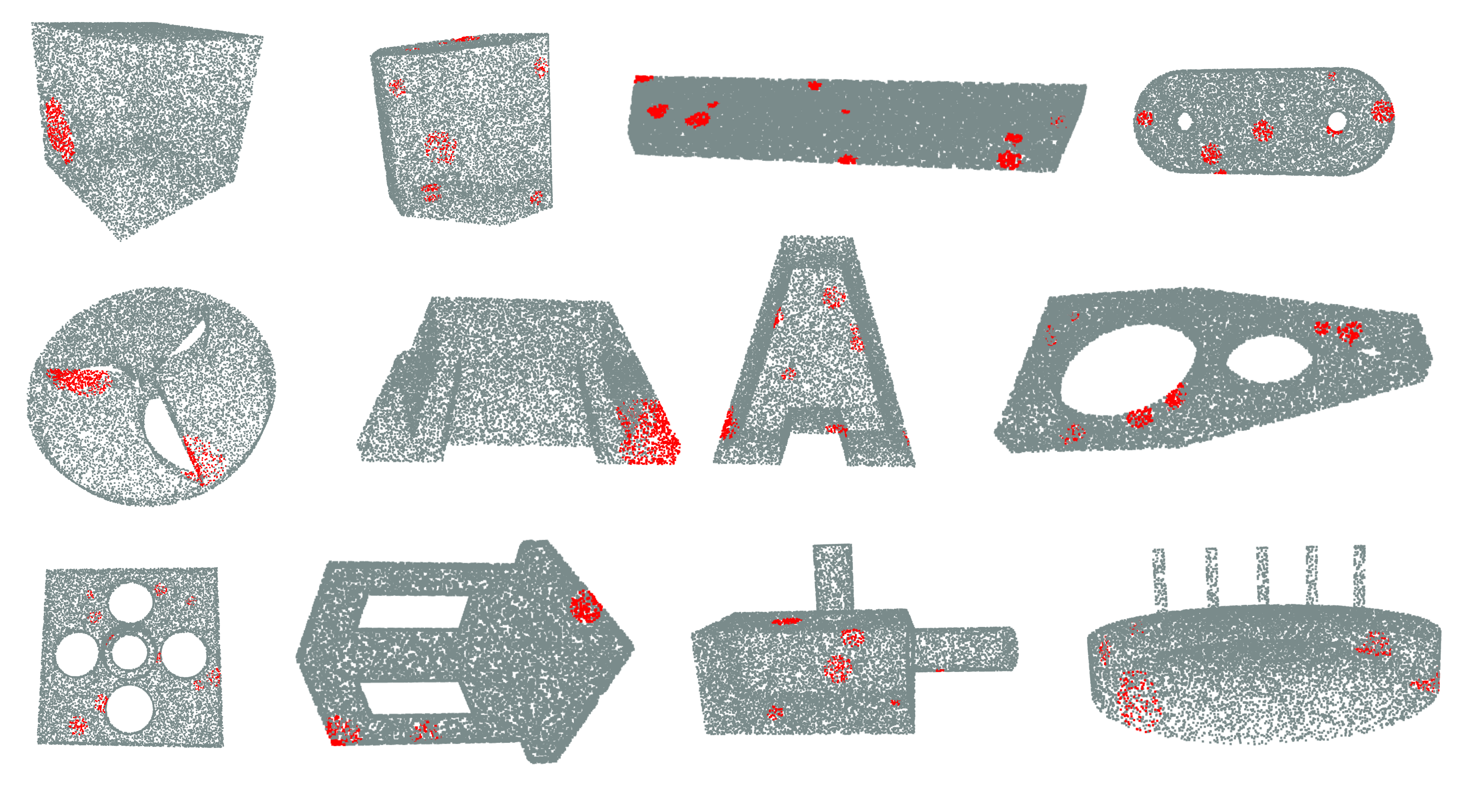}
\end{center}
 \caption{Examples of input point clouds in which holes have been created. The points highlighted in red represent the points that have been removed.}
\label{fig:holes}
\end{figure}

Holes in the input point clouds are created using the following strategy. Firstly, for each point cloud the number holes is selected from a uniform distribution ranging from $1$ to $10$ included. Then, the ratio of points to be removed for each hole is chosen by sampling a normal distribution with mean $0.03$ and standard deviation $0.015$. Finally, for each hole a point is chosen at random and the corresponding number of nearest neighbors points are removed. The nearest neighbors are identified using a geodesic distance computed on the mesh surface. We ensure that the remaining number of points is at least $n=4096$, which corresponds to the number of points used as input. Fig.~\ref{fig:holes} shows some examples of point clouds on which holes have been created.

\subsection{Qualitative Results}

\begin{figure}[ht]
\begin{center}
\includegraphics[width=0.8\linewidth]{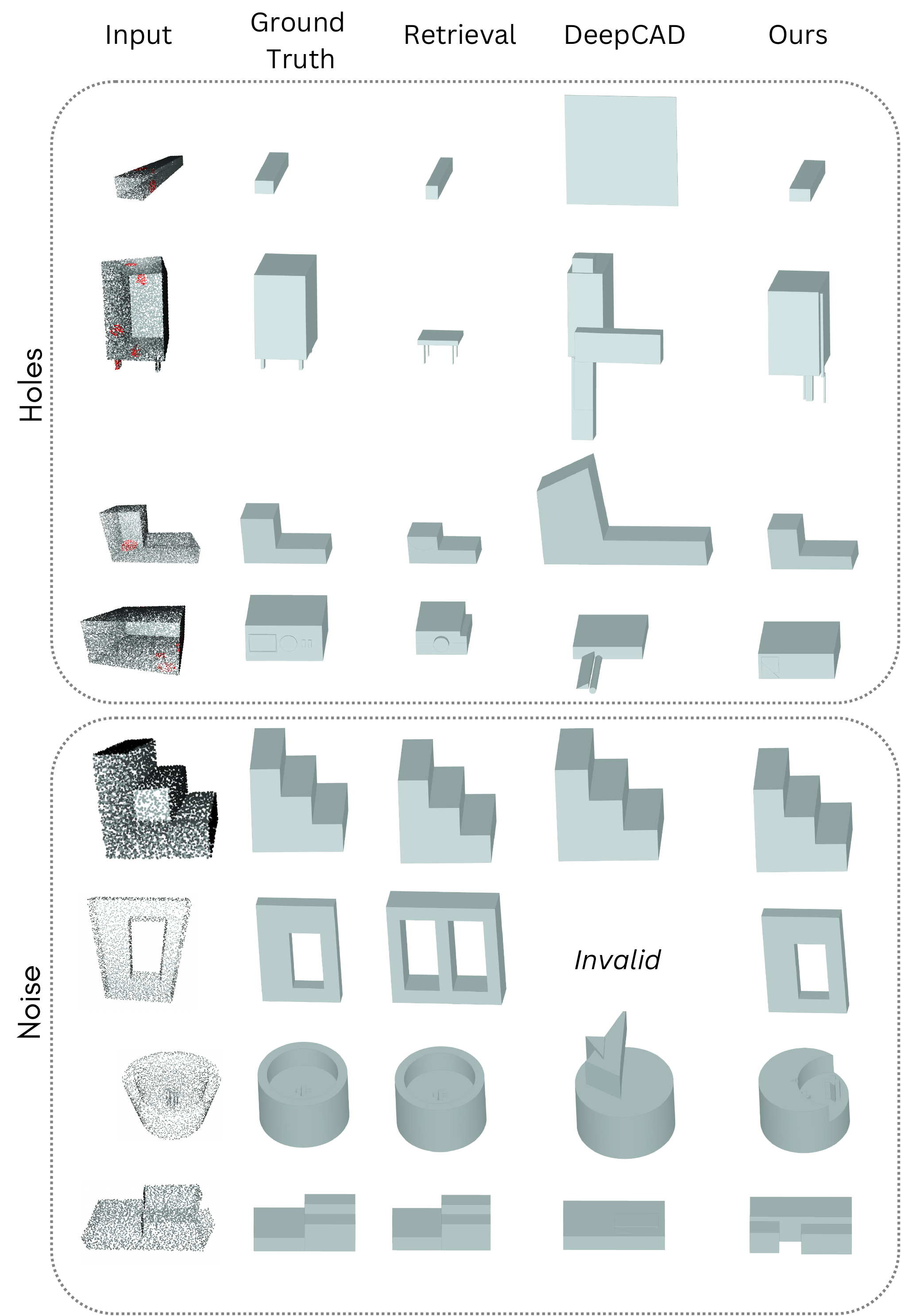}
\end{center}
 \caption{Qualitative results on DeepCAD dataset~\cite{wu2021deepcad} using perturbed input point clouds. The top panel shows results for which holes are created on the input point cloud and the bottom panel shows results for which Perlin noise was applied.}
\label{fig:quali_per}
\end{figure}

Qualitative results for both the hole and noise input perturbations can be found in Fig.~\ref{fig:quali_per}. Those results complement the quantitative results found in Section~\ref{sec:exp_results} of the main paper.

\subsection{Further Quantitative Results}
In this section, further quantitative results for different amounts of point cloud perturbation are presented.
The maximum magnitude of the Perlin noise on the input point cloud is increased compared to the one reported in the main paper (now referred to as \textit{Original Noise}). The results presented in Table~\ref{table:noise_level} show that both \modelname~and DeepCAD are sensitive to the amount of noise. We also perform $2$-epoch finetuning of both models on \textit{Original Noise} training data and report the results (\textit{Finetune}, last row of Table~\ref{table:noise_level}). Notably, \modelname~almost recovers its performance with finetuning.

\begingroup
\setlength{\tabcolsep}{3pt}
\begin{table}[h!]
\centering
\begin{tabular}{c|cc|cc}
\hline
             Noise                                       & \multicolumn{2}{c|}{\modelname} & \multicolumn{2}{c}{DeepCAD~\cite{wu2021deepcad}}\\ 
     increase                    & $APCS\uparrow$ & $CD\downarrow$  & $APCS\uparrow$ & $CD\downarrow$  \\ 
 \hline 
     \textit{Original Noise}           & 0.604 & 18.1 & 0.550 & 31.1\\           
 $+25\%$              & 0.570 & 21.1 & 0.541 & 31.6 \\       
 $+50\%$ &  0.532 & 28.8 & 0.528 & 35.6 \\                 
 \textit{Finetune}            &  0.724 & 5.1 & 0.572  & 27.7        
                           \\ \hline
     
\end{tabular}
\caption{Results on the $APCS$ and $CD$ metrics for different amount of noise added to the input point cloud on the DeepCAD dataset~\cite{wu2021deepcad}.}
\label{table:noise_level}
\end{table}
\endgroup

Furthermore, we conduct an experiment to evaluate the effect of the input cloud sparsity on the performance. Using \modelname~and DeepCAD~\cite{wu2021deepcad} trained with input point clouds of size 4096 points, predictions for the test set with decreasing input point cloud sizes are generated. The results demonstrate that \modelname~is more robust than DeepCAD~\cite{wu2021deepcad} \wrt input sparsity (see Table~\ref{table:input_size}).

\begingroup
\setlength{\tabcolsep}{3pt}
\begin{table}[h!]
\centering
\begin{tabular}{c|cc|cc}
\hline
             Input size                                       & \multicolumn{2}{c|}{\modelname} & \multicolumn{2}{c}{DeepCAD[40]}\\ 
     (\# Points)                    & $APCS\uparrow$ & $CD\downarrow$  & $APCS\uparrow$ & $CD\downarrow$  \\ 
 \hline 
 4096              & 0.732  & 4.5  & 0.604 & 19.2 \\                                  
 2048               & 0.729  & 4.9    &  0.577 & 24.8 \\        
512               & 0.705  & 8.1   & 0.526 & 38.5 \\      
256               &  0.673 & 12.7  & 0.443 & 70.3       
\\ \hline 
\end{tabular}
\caption{Results on the $APCS$ and $CD$ metrics for different size of input clouds on the DeepCAD dataset~\cite{wu2021deepcad}.}
\label{table:input_size}
\end{table}
\endgroup

\clearpage

\end{document}